\RequirePackage{fix-cm}
\documentclass[smallextended]{svjour3}       % onecolumn (second format)
\smartqed  % flush right qed marks, e.g. at end of proof
\usepackage{amsmath}
\usepackage{amssymb,amsfonts}
\usepackage{dsfont}
\usepackage{subfigure}
\usepackage{indentfirst}
\usepackage{mathrsfs}
\usepackage{color}
\usepackage{graphicx}
\usepackage{hyperref}
\usepackage[ruled,linesnumbered]{algorithm2e} % use Times fonts if available on your TeX system
\newcommand{\fR}{\mathcal{R}}
\newcommand{\fO}{\mathcal{O}}
\newcommand{\fN}{\mathcal{N}}
\newcommand{\sR}{\mathbb{R}}
\newcommand{\sN}{\mathbb{N}}

\newcommand{\fE}{\mathcal{E}}
\newtheorem{assumption}{Assumption}
\begin{document}

\title{Homotopy Relaxation Training Algorithms for Infinite-Width Two-Layer ReLU Neural Networks \thanks{YY and WH are supported by National Institute of General Medical Sciences through grant 1R35GM146894.}
}

\titlerunning{HRTA for Infinite-width Two-layer ReLU NNs}        % if too long for running head

\author{Yahong Yang\and Qipin Chen \and Wenrui Hao}

\authorrunning{Y. Yang, Q. Chen, and W. Hao} % if too long for running head

\institute{Yahong Yang \at
              Department of Mathematics,
	The Pennsylvania State University, University Park, State College, PA 16802, USA \\
              \email{yxy5498@psu.edu}           %  \\
           \and
           Qipin Chen \at
              Amazon Prime Video, Seattle, MA 98109, USA\\
	\email{qipinche@amazon.com}\and Wenrui Hao \at
              Department of Mathematics,
	The Pennsylvania State University, University Park, State College, PA 16802, USA \\
              \email{wxh64@psu.edu} 
}

\date{Received: date / Accepted: date}

\maketitle

\begin{abstract}
In this paper, we present a novel training approach called the Homotopy Relaxation Training Algorithm (HRTA), aimed at accelerating the training process in contrast to traditional methods. Our algorithm incorporates two key mechanisms: one involves building a homotopy activation function that seamlessly connects the linear activation function with the ReLU activation function; the other technique entails relaxing the homotopy parameter to enhance the training refinement process. We have conducted an in-depth analysis of this novel method within the context of the neural tangent kernel (NTK), revealing significantly improved convergence rates. Our experimental results, especially when considering networks with larger widths, validate the theoretical conclusions. This proposed HRTA exhibits the potential for other activation functions and deep neural networks.
\end{abstract}

\section{Introduction}

Neural networks (NNs) with the rectified linear unit (ReLU) activation function \cite{glorot2011deep,arora2016understanding} have become increasingly popular in scientific and engineering applications, such as image classification \cite{krizhevsky2017imagenet,he2015delving}, regularization \cite{he2016deep,wu2018deep}. Finding an efficient way to train and obtain the parameters in NNs is an important task, enabling the application of NNs in various domains.

Numerous studies have delved into training methods for NNs, as evidenced by the works of \cite{erhan2010does,keskar2016large,you2019large,he2023side,lu2019scaling,lu2020mean,kingma2014adam,wang2020stochastic,whiting2023convergence,siegel2021accelerated,siegel2021greedy,sun2022adaptive}. However, the optimization of loss functions can become increasingly challenging over time, primarily due to the nonconvex energy landscape. Traditional algorithms such as the gradient descent method and the Adam method often lead to parameter entrapment in local minima or saddle points for prolonged periods. The homotopy training algorithm (HTA) was introduced as a remedy by making slight modifications to the NN structure. HTA draws its roots from the concept of homotopy continuation methods \cite{morgan1987computing,sommese2005numerical,hao2018homotopy,hao2022adaptive}, with its initial introduction found in \cite{chen2019homotopy}. However, constructing a homotopy function requires it to be aligned with the structure of neural networks and entails time-consuming training.

In this paper, we introduce an innovative training approach called the homotopy relaxation training algorithm (HRTA). This approach leverages the homotopy concept, specifically focusing on the activation function, to address the challenges posed by the HRTA. We develop a homotopy activation function that establishes a connection between the linear activation function and the target activation function. By gradually adjusting the homotopy parameter, we enable a seamless transition toward the target activation function. Mathematically, the homotopy activation function is defined as $\sigma_s$, where $s$ is the homotopy parameter, and it takes the form $\sigma_s(x) = (1-s)\text{Id}(x) + s\sigma(x)$. Here, $\text{Id}(x)$ represents the identity function (i.e., the linear activation function), and $\sigma(x)$ is the target activation function.
The term ``homotopy" in the algorithm  signifies its evolution from an entirely linear neural network, where the initial activation function is the identity function ($s=0$). The homotopy activation function undergoes gradual adjustments until it transforms into a target neural network ($s=1$). This transition, from the identity function to the target function, mirrors the principles of homotopy methods. Furthermore, our analysis reveals that by extrapolating (or over-relaxing) the homotopy parameter ($1 < s < 2$), training performance can be further enhanced. In this context, we extend the concept of homotopy training, introducing what we refer to as ``homotopy relaxation training."

In this paper, we relax the homotopy parameter and allow $s$ to take on any positive value in $[0,2]$, rather than being restricted to values in $[0,1]$. Moreover, we provide theoretical support for this algorithm, particularly in hyperparameter scenarios. Our analysis is closely related to neural tangent kernel methods \cite{jacot2018neural,arora2019exact,zhang2020type,cao2020generalization,du2018gradient,huang2023analyzing,chen2020generalized,yang2021tensor,seleznova2022neural,gao2021global,allen2019convergence}. In contrast to classical neural tangent kernel methods, our algorithm studies training performance with changing activation functions during training. In other words, while classical neural tangent kernel methods focus on single neural networks, our algorithm allows for changes in the structure of the neural network due to changes in the activation function during the training. There are other works that introduce parameters during training. They learn the activation during training and identify an adaptive activation function, as demonstrated in \cite{jagtap2020adaptive,jagtap2020locally,jagtap2022deep,agostinelli2014learning}. However, these methods differ from ours. In their approach, they find the proper activation during training, implying a single iteration. The loss function can only decay in one iteration, without another chance to decay. The concept of homotopy in our paper lies in not needing to learn the active function explicitly; instead, we propose a method to alter the function. The loss function has multiple opportunities to decay as we modify the energy landscape between different iterations. We establish that modifying the homotopy parameter at each step increases the smallest eigenvalue of the gradient descent kernel for infinite-width neural networks (see Theorem \ref{large}). Consequently, we present Theorem \ref{convergence} to demonstrate the capacity of HRTA to enhance training speed.

The paper is organized as follows: We introduce the HRTA in Section \ref{HRTA}. Next, in Section \ref{theorem}, we provide theoretical support for our theory. Finally, in Section \ref{experiment}, we conduct experiments, including supervised learning and solving partial differential equations, based on our algorithm.

\section{Homotopy Relaxation Training Algorithm}\label{HRTA}
In this paper, we consider supervised learning for NNs. Within a conventional supervised learning framework, the primary objective revolves around acquiring an understanding of a high dimensional target function denoted as $f(\boldsymbol{x})$ in $[0,1]^d$, with $\|f\|_{L_\infty([0,1]^d)}\le 1$, through a finite collection of data samples $\{(\boldsymbol{x}_i,f(\boldsymbol{x}_i))\}_{i=1}^n$. When embarking on the training of a NN, our aim rests upon the discovery of a NN representation denoted as $\phi(\boldsymbol{x};\boldsymbol{\theta})$ that serves as an approximation to $f(\boldsymbol{x})$, a feat achieved through the utilization of randomly gathered data samples $\{(\boldsymbol{x}_i,f(\boldsymbol{x}_i))\}_{i=1}^n$, with $\boldsymbol{\theta}$ representing the parameters within the NN architecture. It is assumed, in this paper, that the sequence $\{\boldsymbol{x}_i\}_{i=1}^n$ constitutes an independent and identically distributed (i.i.d.) sequence of random variables, uniformly distributed across $[0,1]^d$. Denote 
\begin{align}	\boldsymbol{\theta}_S&:=\arg\min_{\boldsymbol{\theta}}\fR_{S}(\boldsymbol{\theta}):=\arg\min_{\boldsymbol{\theta}}\frac{1}{2n}\sum_{i=1}^n|f(\boldsymbol{x}_i)-\phi(\boldsymbol{x}_i;\boldsymbol{\theta})|^2.\label{thetaS}
\end{align}
Next, we introduce the HRTA by defining $\sigma_{s_p}(x)=(1-s_p)\text{Id}(x)+s_p\sigma(x)$ with a discretized set points of the homotopy parameter $\{s_p\}_{p=1}^M\subset(0,2)$. We then proceed to obtain:
\begin{align}
\boldsymbol{\theta}_S^{s_p} &:= \arg\min_{\boldsymbol{\theta}} \fR_{S, s_p}(\boldsymbol{\theta}), \text{ with an initial guess } \boldsymbol{\theta}_S^{s_{p-1}},p=1,\cdots,M,
\end{align}
where we initialize $\boldsymbol{\theta}_S^{s_{0}}$ randomly, and $\boldsymbol{\theta}_S^{s_M}$ represents the optimal parameter value that we ultimately achieve.

In this paper, we consider a two-layer NN defined as follows \begin{equation}
    \phi(\boldsymbol{x};\boldsymbol{\theta}):=\frac{1}{\sqrt{m}}\sum_{k=1}^m a_k\sigma (\boldsymbol{\omega}_k^{\mathrm{T}}\boldsymbol{x}),
\end{equation} with the activation function $\sigma(z)=\text{ReLU}(z)=\max\{z,0\}$. The evolution of the traditional training can be written as the following differential equation:\begin{equation}
    \frac{\mathrm{d}\boldsymbol{\theta}(t)}{\mathrm{d}t}=-\nabla_{\boldsymbol{\theta}} { \fR_{S}(\boldsymbol{\theta}(t))}.
\end{equation}

In the HRTA setup, we train a sequences of leaky ReLU activate functions \cite{xu2020reluplex,mastromichalakis2020alrelu}. Subsequently, for each of these Leaky ReLUs with given $s_p$,  we train the neural network on a time interval of $[t_{p-1},t_{p}]$:
\begin{equation}
\frac{\mathrm{d}\boldsymbol{\theta}(t)}{\mathrm{d}t}=-\nabla_{\boldsymbol{\theta}} {\fR_{S,s_{p}}(\boldsymbol{\theta}(t))}.
\end{equation}
Moreover, we have $t_0=0$ and initialize the parameter vector $\boldsymbol{\theta}(0)$, drawn from a normal distribution $\fN(\boldsymbol{0},\boldsymbol{I})$. Therefore the HRTA algorithm's progression is outlined in {\bf Algorithm \ref{Alg: RFM}}.

%\begin{remark}\label{speed}
%In the above algorithm, we do not change the width for each iteration. In practical scenarios, we can add some new nodes in each iteration. This approach may help accelerate the training process. The parameters of these new nodes can be selected from normal distributions. Later on, we can provide theoretical support to explain why this can expedite the training.
%\end{remark}
\begin{remark}
    If $s_M=1$, then upon completion, we will have obtained $\boldsymbol{\theta}(t_M)$ and the NN $\phi_{s_M}(\boldsymbol{x};\boldsymbol{\theta}(t_M))$, characterized by pure ReLU activations.
The crux of this algorithm is its ability to transition from Leaky ReLUs to a final configuration of a NN with pure ReLU activations. This transformation is orchestrated via a series of training iterations utilizing the homotopy approach.

However, our paper demonstrates that there is no strict necessity to achieve $s_M=1$. What we aim for is to obtain a NN with parameters $\boldsymbol{\theta}$ that minimizes $\fR_{S,s_M}(\boldsymbol{\theta)}$. This is because, for any value of $s$, we can readily represent $\phi_{s}(\boldsymbol{x};\boldsymbol{\theta})$ as a pure ReLU NN, as shown in the following equation:
\begin{align}\sigma_{s}(x)&=(1-s)\text{Id}(x)+s\sigma(x)\notag\\&=(1-s)\sigma(x)-(1-s)\sigma(-x)+s\sigma(x)=\sigma(x)-(1-s)\sigma(-x).\notag\end{align} To put it simply, if we can effectively train a NN with Leaky-ReLU to learn the target functions, it implies that we can achieve the same level of performance with a NN using standard ReLU activation. Consequently, the theoretical analysis in the paper does not require that the final value of $s_M$ must be set to 1. Moreover, our method is applicable even when $s_M>1$, which we refer to as the relaxation part of HRTA. It's important to highlight that for $s>1$ the decay speed may surpass that of a pure ReLU neural network. This is consistent with the training using the hat activation function (specifically, when $s=2$ in the homotopy activation function) discussed in \cite{hong2022activation}. , although it's worth noting that their work primarily focuses on the linear case (involving only the constant factor change), whereas our work extends this consideration to neural networks.
\end{remark}
\begin{algorithm}\label{Alg: RFM}
\caption{The Homotopy Training Algorithm for Two Layer Neural Networks}
\SetKwData{Left}{left}\SetKwData{This}{this}\SetKwData{Up}{up}
\SetKwFunction{Union}{Union}\SetKwFunction{FindCompress}{FindCompress}
\SetKwInOut{Input}{\textbf{input}}\SetKwInOut{Output}{\textbf{output}}
\Input{Sample points of function $\{(\boldsymbol{x}_i,f(\boldsymbol{x}_i))\}_{i=1}^n$; Initialized homotopy parameter $s_1> 0$; Number of the iteration times $M$;  $\zeta>0$; Training time of each iteration $T$; $t_p=pT$; learning rate $\tau$; %Two layer NNs $\phi_{s_p}(\boldsymbol{x};\boldsymbol{\theta}_0)=\frac{1}{\sqrt{m}}\sum_{k=1}^m a_k\sigma_{s_p} (\boldsymbol{\omega}_k^{\mathrm{T}}\boldsymbol{x})$ with 
$\boldsymbol{\theta}_0\sim \fN(0,\boldsymbol{I})$.}
%\For {$p=1,2,\ldots,M$}{{\boldsymbol{\theta}_{t+1}=\boldsymbol{\theta}_t-\tau \nabla_{\boldsymbol{\theta}}(\fR_{s_p}(\boldsymbol{\theta}_t)}}

\For{$p=1,2,\ldots,M$}{
    \For{$t\in[t_{p-1},t_p]$}{
      $\boldsymbol{\theta}_{t+\tau}=\boldsymbol{\theta}_t-\tau \nabla_{\boldsymbol{\theta}}(\fR_{s_p}(\boldsymbol{\theta}_t))$\;
      %\Up$\leftarrow$ \FindCompress{$Im[i-1,]$}\;
      %\This$\leftarrow$ \FindCompress{$Im[i,j]$}\;
      %\If(\tcp*[h]{O(\Left,\This)==1}){\Left compatible with \This}{\label{lt}
       % \lIf{\Left $<$ \This}{\Union{\Left,\This}}
        %\lElse{\Union{\This,\Left}}
      %}
      %\If(\tcp*[f]{O(\Up,\This)==1}){\Up compatible with \This}{\label{ut}
       % \lIf{\Up $<$ \This}{\Union{\Up,\This}}
        %\tcp{\This is put under \Up to keep tree as flat as possible}\label{cmt}
        %\lElse{\Union{\This,\Up}}\tcp*[h]{\This linked to \Up}\label{lelse}
      %}
    }
    $s_{p+1} := s_p+\zeta$\;
    %\lForEach{element $e$ of the line $i$}{\FindCompress{p}}
  }

\Output {$\phi_{s_M}(\boldsymbol{x},\boldsymbol{\theta}_{T_M})$ as the two layer NN approximation to approximate $f(\boldsymbol{x})$.}
\end{algorithm}

\section{Convergence Analysis}\label{theorem}
\begin{figure}[h!]
		\centering
		\includegraphics[scale=0.30]{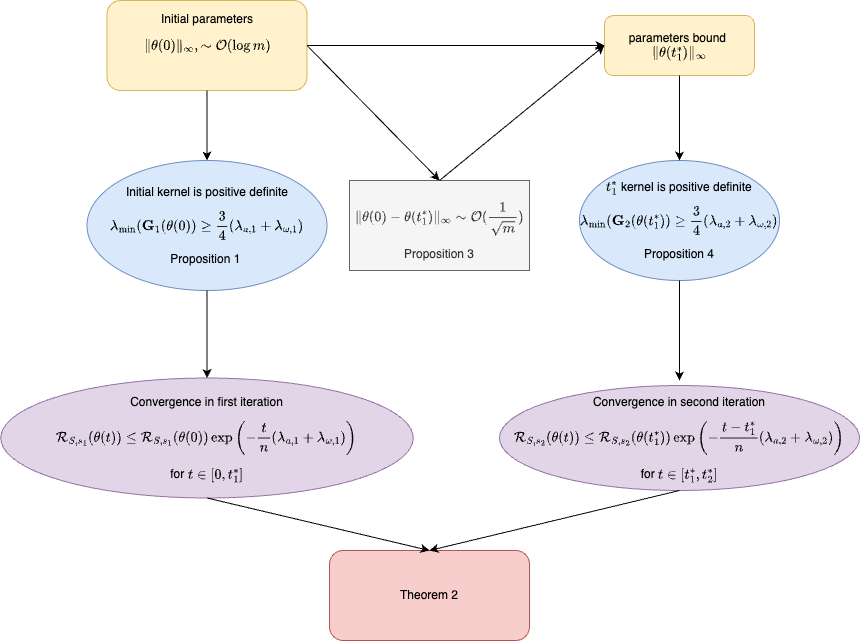}
		\caption{Structure of proof of Theorem \ref{convergence}}
		\label{structure}
	\end{figure}

In this section, we will delve into the convergence analysis of the HRTA based on the neural target kernel methods \cite{jacot2018neural,arora2019exact,zhang2020type,cao2020generalization,du2018gradient,huang2023analyzing,chen2020generalized,yang2021tensor,seleznova2022neural,gao2021global,allen2019convergence}. For simplicity, we will initially focus on the case where $M=2$. Note that for cases with $M>2$, all analyses presented here can be readily extended by repeating the analysis from the first iteration to the second iteration. To start, we set the initial value of $s_1>0$. The structure of the proof of Theorem \ref{convergence} is shown in Figure \ref{structure}.

\subsection{Preliminaries}\label{preliminaries}
		\subsubsection{Neural networks}
		Let us summarize all basic notations used in the NNs as follows:
		
		\textbf{1}. Matrices are denoted by bold uppercase letters. For example, $\boldsymbol{A}\in\sR^{m\times n}$ is a real matrix of size $m\times n$ and $\boldsymbol{A}^{\mathrm{T}}$ denotes the transpose of $\boldsymbol{A}$. $\|\boldsymbol{A}\|_F$ is the Frobenius norm of the matrix $\boldsymbol{A}$.
		
		\textbf{2}. Vectors are denoted by bold lowercase letters. For example, $\boldsymbol{v}\in\sR^n$ is a column vector of size $n$.

		%\textbf{4}. Let $B_{r,|\cdot|}(\boldsymbol{x})\subset\sR^d$ be the closed ball with a center $\boldsymbol{x}\in\sR^d$ and a radius $r$ measured by the Euclidean distance. Similarly, $B_{r,\|\cdot\|_{\ell_\infty}}(\boldsymbol{x})\subset\sR^d$ be the closed ball with a center $\boldsymbol{x}\in\sR^d$ and a radius $r$ measured by the $\ell_\infty$-norm.
		
		\textbf{3}. Assume $\boldsymbol{n}\in\sN_+^n$, then $f(\boldsymbol{n})=\fO(g(\boldsymbol{n}))$ means that there exists positive $C$ independent of $\boldsymbol{n},f,g$ such that $f(\boldsymbol{n})\le Cg(\boldsymbol{n})$ when all entries of $\boldsymbol{n}$ go to $+\infty$.
		
		\textbf{4}. Recall $\sigma(x)=\max\{0,x\}$ and $\sigma_{s}(x)=(1-s)\text{Id}(x)+s\sigma(x)$ for $s>0$. Two-layer NN structures are defined by:\begin{equation}
    \phi_{s_p}(\boldsymbol{x};\boldsymbol{\theta}):=\frac{1}{\sqrt{m}}\sum_{k=1}^m a_k\sigma_{s_p} (\boldsymbol{\omega}_k^{\mathrm{T}}\boldsymbol{x}).
\end{equation}

  \textbf{5}. Recall that \begin{equation}
    \fR_{S,s_p}(\boldsymbol{\theta}):=\frac{1}{2n}\sum_{i=1}^n|f(\boldsymbol{x}_i)-\phi_{s_p}(\boldsymbol{x}_i;\boldsymbol{\theta})|^2,
\end{equation} it is assumed that the sequence $\{\boldsymbol{x}_i\}_{i=1}^n$ consists of independent and identically distributed (i.i.d.) random variables. These random variables are uniformly distributed within the hypercube $[0,1]^d$, where $d$ is the dimension of the input space.
\subsection{Gradient descent kernel}
The kernels characterizing the training dynamics for the $p$-th iteration take the following form:\begin{align}
    k^{[a]}_p(\boldsymbol{x},\boldsymbol{x}'):=&\mathbf{E}_{\boldsymbol{\omega}}\sigma_{s_p}(\boldsymbol{\omega}^{\mathrm{T}}\boldsymbol{x})\sigma_{s_p}(\boldsymbol{\omega}^{\mathrm{T}}\boldsymbol{x}')\notag\\k^{[\boldsymbol{\omega}]}_p(\boldsymbol{x},\boldsymbol{x}'):=&\mathbf{E}_{(a,\boldsymbol{\omega})}a^2\sigma'_{s_p}(\boldsymbol{\omega}^{\mathrm{T}}\boldsymbol{x})\sigma'_{s_p}(\boldsymbol{\omega}^{\mathrm{T}}\boldsymbol{x}')\boldsymbol{x}\cdot\boldsymbol{x}'.
\end{align} The Gram matrices, denoted as $\boldsymbol{K}^{[a]}_p$ and $\boldsymbol{K}^{[\boldsymbol{\omega}]}_p$, corresponding to an infinite-width two-layer network with the activation function $\sigma_{s_p}$, can be expressed as follows:\begin{align}
    &K_{ij,p}^{[a]}=k^{[a]}_p(\boldsymbol{x}_i,\boldsymbol{x}_j),~\boldsymbol{K}^{[a]}_p=(K_{ij,p}^{[a]})_{n\times n},\notag\\& K_{ij,p}^{[\boldsymbol{\omega}]}=k^{[\boldsymbol{\omega}]}_p(\boldsymbol{x}_i,\boldsymbol{x}_j),~\boldsymbol{K}^{[\boldsymbol{\omega}]}_p=(K_{ij,p}^{[\boldsymbol{\omega}]})_{n\times n}
\end{align} Moreover, the Gram matrices, referred to as $\boldsymbol{G}^{[a]}_p$ and $\boldsymbol{G}^{[\boldsymbol{\omega}]}_p$, corresponding to a finite-width two-layer network with the activation function $\sigma_{s_p}$, can be defined as\begin{align}
    &G_{ij,p}^{[a]}=\frac{1}{m}\sum_{k=1}^{m}\sigma_{s_p}(\boldsymbol{\omega}^{\mathrm{T}}_k\boldsymbol{x})\sigma_{s_p}(\boldsymbol{\omega}^{\mathrm{T}}_k\boldsymbol{x}'),~\boldsymbol{G}^{[a]}_p=(G_{ij,p}^{[a]})_{n\times n},\notag\\& G_{ij,p}^{[\boldsymbol{\omega}]}=\frac{1}{m}\sum_{k=1}^{m}a^2\sigma'_{s_p}(\boldsymbol{\omega}^{\mathrm{T}}_k\boldsymbol{x})\sigma'_{s_p}(\boldsymbol{\omega}^{\mathrm{T}}_k\boldsymbol{x}')\boldsymbol{x}\cdot\boldsymbol{x}',~\boldsymbol{K}^{[\boldsymbol{\omega}]}_p=(G_{ij,p}^{[\boldsymbol{\omega}]})_{n\times n}.
\end{align}

%Denote $\boldsymbol{K}^{[a]}$ and $\boldsymbol{K}^{[\boldsymbol{\omega}]}$ are the Gram matrices and 

\begin{assumption}\label{positive} Denote $\boldsymbol{K}^{[a]}$ and $\boldsymbol{K}^{[\boldsymbol{\omega}]}$ are the Gram matrices for ReLU neural network and 
 \begin{align}
    &H_{ij}^{[a]}=k^{[a]}(-\boldsymbol{x}_i,\boldsymbol{x}_j),~\boldsymbol{H}^{[a]}=(H_{ij}^{[a]})_{n\times n},~~H_{ij}^{[\boldsymbol{\omega}]}=k^{[\boldsymbol{\omega}]}(-\boldsymbol{x}_i,\boldsymbol{x}_j),~\boldsymbol{H}^{[\boldsymbol{\omega}]}=(H_{ij}^{[\boldsymbol{\omega}]})_{n\times n}\notag\\& M_{ij}^{[a]}=k^{[a]}(\boldsymbol{x}_i,-\boldsymbol{x}_j),~\boldsymbol{M}^{[a]}=(M_{ij}^{[a]})_{n\times n},~~M_{ij}^{[\boldsymbol{\omega}]}=k^{[\boldsymbol{\omega}]}(\boldsymbol{x}_i,-\boldsymbol{x}_j),~\boldsymbol{M}^{[\boldsymbol{\omega}]}=(M_{ij}^{[\boldsymbol{\omega}]})_{n\times n}\notag\\&T_{ij}^{[a]}=k^{[a]}(-\boldsymbol{x}_i,-\boldsymbol{x}_j),~\boldsymbol{T}^{[a]}=(T_{ij}^{[a]})_{n\times n},~~T_{ij}^{[\boldsymbol{\omega}]}=k^{[\boldsymbol{\omega}]}(-\boldsymbol{x}_i,-\boldsymbol{x}_j),~\boldsymbol{T}^{[\boldsymbol{\omega}]}=(T_{ij}^{[\boldsymbol{\omega}]})_{n\times n}.\notag\end{align}Suppose that all matrices defined above are strictly positive definite.
    %In other words,\begin{align}
        %\lambda_{a,1}:=\lambda_{\text{min}}\left(\boldsymbol{K}^{[a]}_1\right)>0,~~ \lambda_{\boldsymbol{\omega},1}:=\lambda_{\text{min}}\left(\boldsymbol{K}^{[\boldsymbol{\omega}]}_1\right)>0.
    %\end{align}
\end{assumption}
\begin{remark}
   We would like to point out that if, for all $i$ and $j$ satisfying $i \neq j$, we have $\pm \boldsymbol{x}_i$ not parallel to $\pm \boldsymbol{x}_j$, then Assumption \ref{positive} is satisfied. The validity of this assertion can be established by referring to \cite[Theorem 3.1]{du2018gradient} for the proof.\end{remark}

\begin{theorem}\label{large}
    Suppose Assumption \ref{positive} holds, denote \[
        \lambda_{a,p}:=\lambda_{\text{min}}\left(\boldsymbol{K}^{[a]}_p\right),~~ \lambda_{\boldsymbol{\omega},p}:=\lambda_{\text{min}}\left(\boldsymbol{K}^{[\boldsymbol{\omega}]}_p\right).
    \]Then we have \(\lambda_{\boldsymbol{\omega},p+1}\ge\lambda_{\boldsymbol{\omega},p}>0,~\lambda_{a,p+1}\ge\lambda_{a,p}>0
    \) for all $0\le s_p\le s_{p+1}$.
\end{theorem}
Before the proof, we need a lemma in the linear algebra.
\begin{lemma}[Weyl’s Inequalities]\label{eig}
    Suppose $\boldsymbol{A}$ and $\boldsymbol{B} $ are real symmetric matrixs, we have that\begin{align}
        \lambda_{\text{min}}(\boldsymbol{A}+\boldsymbol{B}) \geq \lambda_{\text{min}}(\boldsymbol{A})+ \lambda_{\text{min}}(\boldsymbol{B}).
    \end{align}
\end{lemma}
% \begin{proof}
%    Let $\lambda_a$ be defined as $\lambda_{\text{min}}(\boldsymbol{A})$ and $\lambda_b$ as $\lambda_{\text{min}}(\boldsymbol{B})$. Consequently, we can assert that $\boldsymbol{A}+\boldsymbol{B}-(\lambda_a+\lambda_b)\boldsymbol{I}$ possesses positive definiteness. If we designate $\lambda$ as an eigenvalue of $\boldsymbol{A}+\boldsymbol{B}$, then it follows that $(\boldsymbol{A}+\boldsymbol{B})\boldsymbol{x}_*=\lambda \boldsymbol{x}_*$. This relationship can be expressed as:

% \begin{equation}
% (\boldsymbol{A}+\boldsymbol{B}-(\lambda_a+\lambda_b)\boldsymbol{I})\boldsymbol{x}_*=(\lambda-\lambda_a-\lambda_b)\boldsymbol{x}_*.
% \end{equation}

% Consequently, we can deduce that $\lambda \geq \lambda_a+\lambda_b$, which further implies that $\lambda_{\text{min}}(\boldsymbol{A}+\boldsymbol{B}) \geq \lambda_{\text{min}}(\boldsymbol{A})+ \lambda_{\text{min}}(\boldsymbol{B})$.
% \end{proof}

\begin{proof}[{Proof of Theorem \ref{large}}]
    For the case $s_p<1$, let's start by considering the expression for the matrix $\boldsymbol{K}^{[\boldsymbol{\omega}]}_p$ where \begin{align}
        \boldsymbol{K}^{[\boldsymbol{\omega}]}_p=(K_{ij,p}^{[\boldsymbol{\omega}]})_{n\times n}=\left(\mathbf{E}_{(a,\boldsymbol{\omega})}a^2\sigma'_{s_p}(\boldsymbol{\omega}^{\mathrm{T}}\boldsymbol{x}_i)\sigma'_{s_p}(\boldsymbol{\omega}^{\mathrm{T}}\boldsymbol{x}_j)\boldsymbol{x}_i\cdot\boldsymbol{x}_j\right)_{n\times n}.
    \end{align} Given the derivative of the activation function: \[\sigma'_{s_p}(x)=\left\{\begin{array}{l}
1,~x> 0 \\
(1-s_p),~x< 0\\0,~ x=0 \end{array}\right.~~\sigma'_{s_{p+1}}(x)=\left\{\begin{array}{l}
1,~x> 0 \\
(1-s_{p+1}),~x< 0\\0,~ x=0 \end{array}\right.
\] we have \begin{equation}
    \sigma'_{s_{p+1}}(x)=\sigma'_{s_{p}}(x)+(s_p-s_{p+1})\sigma (-x)
\end{equation}\begin{align}
&\mathbf{E}_{(a,\boldsymbol{\omega})}a^2\sigma'_{s_{p+1}}(\boldsymbol{\omega}^{\mathrm{T}}\boldsymbol{x}_i)\sigma'_{s_{p+1}}(\boldsymbol{\omega}^{\mathrm{T}}\boldsymbol{x}_j)\boldsymbol{x}_i\cdot\boldsymbol{x}_j\notag\\=&\mathbf{E}_{(a,\boldsymbol{\omega})}a^2\sigma'_{s_p}(\boldsymbol{\omega}^{\mathrm{T}}\boldsymbol{x}_i)\sigma'_{s_p}(\boldsymbol{\omega}^{\mathrm{T}}\boldsymbol{x}_j)\boldsymbol{x}_i\cdot\boldsymbol{x}_j\notag\\&-(s_p-s_{p+1})\mathbf{E}_{(a,\boldsymbol{\omega})}a^2\big[\sigma'(\boldsymbol{\omega}^{\mathrm{T}}\cdot(-\boldsymbol{x}_i))\sigma'_{s_p}(\boldsymbol{\omega}^{\mathrm{T}}\boldsymbol{x}_j)(-\boldsymbol{x}_i)\cdot\boldsymbol{x}_j\notag\\&+\sigma'_{s_p}(\boldsymbol{\omega}^{\mathrm{T}}\boldsymbol{x}_i)\sigma'(\boldsymbol{\omega}^{\mathrm{T}}\cdot(-\boldsymbol{x}_j))\boldsymbol{x}_i\cdot(-\boldsymbol{x}_j)\big]\notag\\&+(s_p-s_{p+1})^2\mathbf{E}_{(a,\boldsymbol{\omega})}a^2\sigma'(\boldsymbol{\omega}^{\mathrm{T}}\cdot(-\boldsymbol{x}_i))\sigma'(\boldsymbol{\omega}^{\mathrm{T}}\cdot(-\boldsymbol{x}_j))\boldsymbol{x}_i\cdot\boldsymbol{x}_j.
\end{align} Furthermore, since \begin{equation}\sigma'(x)=\frac{\sigma'_{s_p}(x)-s_p\sigma'_{s_p}(-x)}{1-s_p^2},\end{equation} we have \begin{align}
    &\sigma'(\boldsymbol{\omega}^{\mathrm{T}}\cdot(-\boldsymbol{x}_i))\sigma'_{s_p}(\boldsymbol{\omega}^{\mathrm{T}}\boldsymbol{x}_j)(-\boldsymbol{x}_i)\cdot\boldsymbol{x}_j\notag\\
    =&\frac{1}{1-s_p^2}\left[\sigma'_{s_p}(\boldsymbol{\omega}^{\mathrm{T}}(-\boldsymbol{x}_i))\sigma'_{s_p}(\boldsymbol{\omega}^{\mathrm{T}}\boldsymbol{x}_j)(-\boldsymbol{x}_i)\boldsymbol{x}_j+s_p\sigma'_{s_p}(\boldsymbol{\omega}^{\mathrm{T}}\boldsymbol{x}_i)\sigma'_{s_p}(\boldsymbol{\omega}^{\mathrm{T}}\boldsymbol{x}_j)\boldsymbol{x}_i\boldsymbol{x}_j\right]
\notag\\&\sigma'_{s_p}(\boldsymbol{\omega}^{\mathrm{T}}\boldsymbol{x}_i)\sigma'(\boldsymbol{\omega}^{\mathrm{T}}(-\boldsymbol{x}_j))\boldsymbol{x}_i\cdot(-\boldsymbol{x}_j)\notag\\
    =&\frac{1}{1-s_p^2}\left[\sigma'_{s_p}(\boldsymbol{\omega}^{\mathrm{T}}\boldsymbol{x}_i)\sigma'_{s_p}(\boldsymbol{\omega}^{\mathrm{T}}(-\boldsymbol{x}_j))(-\boldsymbol{x}_i)\boldsymbol{x}_j+s_p\sigma'_{s_p}(\boldsymbol{\omega}^{\mathrm{T}}\boldsymbol{x}_i)\sigma'_{s_p}(\boldsymbol{\omega}^{\mathrm{T}}\boldsymbol{x}_j)\boldsymbol{x}_i\boldsymbol{x}_j\right].\end{align}Therefore, \begin{equation}\boldsymbol{K}^{[\boldsymbol{\omega}]}_{p+1}=\left(1+\frac{2s_p(s_{p+1}-s_p)}{1-s_p^2}\right)\boldsymbol{K}^{[\boldsymbol{\omega}]}_p+\frac{s_{p+1}-s_p}{1-s_p^2}(\boldsymbol{M}^{[\boldsymbol{\omega}]}_p+\boldsymbol{H}^{[\boldsymbol{\omega}]}_p)+(s_{p+1}-s_p)^2\boldsymbol{T}^{[\boldsymbol{\omega}]}_M.\label{iteration}\end{equation}
    When $s_p < 1$, with the initial condition $s_0=0$, we can establish the following inequalities based on Assumption \ref{positive}, where $\boldsymbol{K}^{[\boldsymbol{\omega}]}_0, \boldsymbol{M}^{[\boldsymbol{\omega}]}_0, \boldsymbol{H}^{[\boldsymbol{\omega}]}_0$ is positive, and Lemma \ref{eig} holds:
\begin{equation}
\lambda_{\text{min}}(\boldsymbol{K}^{[\boldsymbol{\omega}]}_{1}) \geq 0.
\label{beg}\end{equation}
The reason why $\boldsymbol{K}^{[\boldsymbol{\omega}]}_0, \boldsymbol{M}^{[\boldsymbol{\omega}]}_0, \boldsymbol{H}^{[\boldsymbol{\omega}]}_0$ are positive definite matrices is indeed attributed to the fact that $\sigma'_0(x)$ is a constant function. Specifically, for $\boldsymbol{K}^{[\boldsymbol{\omega}]}_0,$ it can be represented as \[(a(\boldsymbol{x}_1,\boldsymbol{x}_2,\ldots,\boldsymbol{x}_n))^{\mathrm{T}}\cdot a(\boldsymbol{x}_1,\boldsymbol{x}_2,\ldots,\boldsymbol{x}_n),\] which is inherently positive definite. Similar propositions can be derived for $\boldsymbol{M}^{[\boldsymbol{\omega}]}_0$ and $\boldsymbol{H}^{[\boldsymbol{\omega}]}_0$ based on the same principle.

Now, when $0 \leq s_p \leq s_{p+1}$ and $s_p < 1$ and Lemma \ref{eig} holds:
\[
\lambda_{\text{min}}(\boldsymbol{K}^{[\boldsymbol{\omega}]}_{p+1}) \geq \lambda_{\text{min}}(\boldsymbol{K}^{[\boldsymbol{\omega}]}_p) \geq 0
\] due to Eqs.~(\ref{iteration},\ref{beg}).

For the case $s_p\ge1$, we have that\begin{align}
&\mathbf{E}_{(a,\boldsymbol{\omega})}a^2\sigma'_{s_{p+1}}(\boldsymbol{\omega}^{\mathrm{T}}\boldsymbol{x}_i)\sigma'_{s_{p+1}}(\boldsymbol{\omega}^{\mathrm{T}}\boldsymbol{x}_j)\boldsymbol{x}_i\cdot\boldsymbol{x}_j\notag\\=&\mathbf{E}_{(a,\boldsymbol{\omega})}a^2\sigma'_{s_p}(\boldsymbol{\omega}^{\mathrm{T}}\boldsymbol{x}_i)\sigma'_{s_p}(\boldsymbol{\omega}^{\mathrm{T}}\boldsymbol{x}_j)\boldsymbol{x}_i\cdot\boldsymbol{x}_j\notag\\&-(s_p-s_{p+1})\mathbf{E}_{(a,\boldsymbol{\omega})}a^2\big[\sigma'(\boldsymbol{\omega}^{\mathrm{T}}\cdot(-\boldsymbol{x}_i))\sigma'_{s_p}(\boldsymbol{\omega}^{\mathrm{T}}\boldsymbol{x}_j)(-\boldsymbol{x}_i)\cdot\boldsymbol{x}_j\notag\\&+\sigma'_{s_p}(\boldsymbol{\omega}^{\mathrm{T}}\boldsymbol{x}_i)\sigma'(\boldsymbol{\omega}^{\mathrm{T}}\cdot(-\boldsymbol{x}_j))\boldsymbol{x}_i\cdot(-\boldsymbol{x}_j)\big]\notag\\&+(s_p-s_{p+1})^2\mathbf{E}_{(a,\boldsymbol{\omega})}a^2\sigma'(\boldsymbol{\omega}^{\mathrm{T}}\cdot(-\boldsymbol{x}_i))\sigma'(\boldsymbol{\omega}^{\mathrm{T}}\cdot(-\boldsymbol{x}_j))\boldsymbol{x}_i\cdot\boldsymbol{x}_j.
\end{align} Furthermore, since \begin{equation}\sigma'_{s_p}(x)=\sigma'(x)+(1-s_p)\sigma'(-x),\end{equation} we have \begin{align}
&\sigma'(\boldsymbol{\omega}^{\mathrm{T}}\cdot(-\boldsymbol{x}_i))\sigma'_{s_p}(\boldsymbol{\omega}^{\mathrm{T}}\boldsymbol{x}_j)(-\boldsymbol{x}_i)\cdot\boldsymbol{x}_j\notag\\
=&\sigma'(\boldsymbol{\omega}^{\mathrm{T}}\cdot(-\boldsymbol{x}_i))\sigma'(\boldsymbol{\omega}^{\mathrm{T}}\boldsymbol{x}_j)(-\boldsymbol{x}_i)\cdot\boldsymbol{x}_j-(1-s_p)\sigma'(\boldsymbol{\omega}^{\mathrm{T}}(-\boldsymbol{x}_i))\sigma'(\boldsymbol{\omega}^{\mathrm{T}}(-\boldsymbol{x}_j))(-\boldsymbol{x}_i)(-\boldsymbol{x}_j)
\notag\\&\sigma'_{s_p}(\boldsymbol{\omega}^{\mathrm{T}}\boldsymbol{x}_i)\sigma'(\boldsymbol{\omega}^{\mathrm{T}}(-\boldsymbol{x}_j))\boldsymbol{x}_i\cdot(-\boldsymbol{x}_j)\notag\\
=&\sigma'(\boldsymbol{\omega}^{\mathrm{T}}\cdot\boldsymbol{x}_i)\sigma'(\boldsymbol{\omega}^{\mathrm{T}}(-\boldsymbol{x}_j))(-\boldsymbol{x}_i)\cdot\boldsymbol{x}_j-(1-s_p)\sigma'(\boldsymbol{\omega}^{\mathrm{T}}(-\boldsymbol{x}_i))\sigma'(\boldsymbol{\omega}^{\mathrm{T}}(-\boldsymbol{x}_j))(-\boldsymbol{x}_i)(-\boldsymbol{x}_j)\notag.\end{align}Therefore, \[\boldsymbol{K}^{[\boldsymbol{\omega}]}_{p+1}=\boldsymbol{K}^{[\boldsymbol{\omega}]}_p-(1-s_p)(s_{p+1}-s_p)(\boldsymbol{M}^{[\boldsymbol{\omega}]}_M+\boldsymbol{H}^{[\boldsymbol{\omega}]}_M)+(s_{p+1}-s_p)(s_{p+1}-s_p+2)\boldsymbol{T}^{[\boldsymbol{\omega}]}_M.\] When $0\le s_p\le s_{p+1}$ and $s_p\ge1$, we have that \[\lambda_{\text{min}}(\boldsymbol{K}^{[\boldsymbol{\omega}]}_{p+1})\ge \lambda_{\text{min}}(\boldsymbol{K}^{[\boldsymbol{\omega}]}_p)\ge 0\] based on Assumption \ref{positive}, as well as Lemma \ref{eig}. Similar results can be derived for the Gram matrices with respect to the parameter $a$.
\end{proof}

\subsection{Convergence of $t_1$ iteration}

The convergence of the first iteration aligns with the traditional neural target kernel analysis. For brevity, we omit the proof in this subsection; however, readers can find comprehensive details in \cite{jacot2018neural,gao2021global,luo2021phase}. Here, we introduce the three results that will be employed in the subsequent iterations.
\begin{lemma}[bounds of initial parameters \cite{luo2021phase}]\label{bound initial}
    Given $\delta\in(0,1)$, we have with probability at least $1-\delta$ over the choice of $\boldsymbol{\theta}(0)$ such that \begin{equation}
        \max_{k\in[m]}\{|a_k(0)|,\|\boldsymbol{\omega}_k(0)\|_\infty\}\le \sqrt{2\log\frac{2m(d+1)}{\delta}}.
    \end{equation}
\end{lemma}

\begin{proposition}[\cite{luo2021phase}]\label{eig pos}
   Given the sample set $S=\left\{\left(\boldsymbol{x}_i, y_i\right)\right\}_{i=1}^n \subset \Omega$ with $\boldsymbol{x}_i$ 's drawn i.i.d. with uniformly distributed and $\delta \in(0,1)$. Suppose that Assumption \ref{positive} holds. If $m \geq \frac{16 n^2 d^2 C_{\psi, d}}{C_0 \lambda^2} \log \frac{4 n^2}{\delta}$ then with probability at least $1-\delta$ over the choice of $\boldsymbol{\theta}(0)$, we have
$$
\lambda_{\min }\left(\boldsymbol{G}_1\left(\boldsymbol{\theta}(0)\right)\right) \geq\frac{3}{4}(\lambda_{a,1}+\lambda_{\boldsymbol{\omega},1}),
$$where $C_0$ is an absolute constant shown in Proposition \ref{vershynin}.
\end{proposition}

Set \begin{equation}t^*_1=\inf\{t\mid\boldsymbol{\theta}(t)\not\in\fN_1(\boldsymbol{\theta}(0))\}\label{t_1}\end{equation} where \[\fN_1(\boldsymbol{\theta}(0)):=\left\{\boldsymbol{\theta}\mid\|\boldsymbol{G}_2(\boldsymbol{\theta})-\boldsymbol{G}_1(\boldsymbol{\theta}(0))\|_F\le \frac{1}{4}(\lambda_{a,1}+\lambda_{\boldsymbol{\omega},1})\right\}.\]

Note that \( t^*_1 \) represents the maximum time parameter, ensuring that the smallest eigenvalue of the dynamic is not small and positive. This condition facilitates an exponential decay of the loss functions. Outside this designated region, the rate of decay might decelerate, making it challenging to achieve rapid convergence.

The principle behind the homotopy approach involves modifying the activation functions to alter the energy landscape of the loss functions. This adjustment aims to ensure that the smallest eigenvalue remains large and positive. The rationale for this modification stems from the concept of \( \boldsymbol{\theta} \)-lazy training in the neural target kernel and the corresponding increase in the smallest eigenvalue (Theorem \ref{large}). Further details on this topic will be elaborated upon in subsequent sections.

\begin{proposition}[\cite{luo2021phase}]\label{convergence 1}
   Given the sample set $S=\left\{\left(\boldsymbol{x}_i, y_i\right)\right\}_{i=1}^n \subset \Omega$ with $\boldsymbol{x}_i$ 's drawn i.i.d. with uniformly distributed and $\delta \in(0,1)$. Suppose that Assumption \ref{positive} holds. If $m \geq \frac{16 n^2 d^2 C_{\psi, d}}{C_0 \lambda^2} \log \frac{4 n^2}{\delta}$ then with probability at least $1-\delta$ over the choice of $\boldsymbol{\theta}(0)$, we have for any $t\in[0,t^*_1]$
\begin{equation}
    \fR_{S,s_1}(\boldsymbol{\theta}(t))\le \fR_{S,s_1}(\boldsymbol{\theta}(0))\exp\left(-\frac{t}{n} (\lambda_{a,1}+\lambda_{\boldsymbol{\omega},1})\right),
\end{equation}where $C_0$ is an absolute constant shown in Proposition \ref{vershynin}.
\end{proposition}

\subsection{Convergence of $t_2$ iteration}\label{t2 it}
In this paper, without sacrificing generality, we focus our attention on the case where $M=2$. However, it's important to note that our analysis and methodology can readily be extended to the broader scenario of $M\geq 2$. 
%All the proof in this paper can be found in Appendix \ref{proof t_2}.

The following method presents the \( \boldsymbol{\theta} \)-lazy training. This proposition is crucial, as if \( \boldsymbol{\theta}(t^*_1) \) deviates significantly from the initial value, even with the smallest eigenvalue of the dynamical system increasing during the homotopy and relaxation training, the gap may cause the dynamical system at the beginning of the second iteration, i.e., the end of the first iteration, to be not positive.

    \begin{proposition}[{\cite{luo2021phase}}]\label{bound dy}
    Given the sample set $S=\left\{\left(\boldsymbol{x}_i, y_i\right)\right\}_{i=1}^n \subset \Omega$ with $\boldsymbol{x}_i$ 's drawn i.i.d. with uniformly distributed and $\delta \in(0,1)$. Suppose that Assumption \ref{positive} holds. If $$m \geq \max\left\{\frac{16 n^2 d^2 C_{\psi, d}}{C_0 \lambda^2} \log \frac{4 n^2}{\delta},\frac{8n^2d^2  \fR_{S,s_1}\left(\boldsymbol{\theta}(0)\right)}{(\lambda_{a,1}+\lambda_{\boldsymbol{\omega},1})^2}\right\}$$ then with probability at least $1-\delta$ over the choice of $\boldsymbol{\theta}(0)$, we have for any $t\in[0,t^*_1]$
\begin{align}
        &\max_{k\in[m]}\{|a_k(t)-a_k(0)|,\|\boldsymbol{\omega}_k(t)-\boldsymbol{\omega}_k(0)\|_\infty\}\notag\\\le  &\frac{8\sqrt{2}n d  \sqrt{\fR_{S,s_1}\left(\boldsymbol{\theta}(0)\right)}}{\sqrt{m}(\lambda_{a,1}+\lambda_{\boldsymbol{\omega},1})}\sqrt{2\log\frac{4m(d+1)}{\delta}},\notag
    \end{align}where $C_0$ is an absolute constant shown in Proposition \ref{vershynin}.
\end{proposition}

For simplicity, we define a $\fO\left(\frac{\log m}{\sqrt{m}}\right)$ term $\psi$, which is \[\psi(m):=\frac{8\sqrt{2}n d  \sqrt{\fR_{S,s_1}\left(\boldsymbol{\theta}(0)\right)}}{(\lambda_{a,1}+\lambda_{\boldsymbol{\omega},1})}\sqrt{2\log\frac{4m(d+1)}{\delta}}.\]

Moving forward, we will employ $\boldsymbol{\theta}(t_1^*)$ as the initial value for training over $t_2$ iterations. However, before we proceed, it is crucial to carefully select the value of $s_2$. This choice of $s_2$ depends on both $\fR_{S,s_1}(\boldsymbol{\theta}(t_1^*))$ and a constant $\zeta$, with the condition that $\zeta$ is a positive constant, ensuring that $0 < \zeta$. Therefore, we define $s_2$ as:
\(
s_2 = s_1 + \zeta
\)
where $\zeta > 0$ is a constant.

It's important to emphasize that for each $\boldsymbol{\theta}(t_1^*)$, given that the training dynamics system operates without any random elements, we can determine it once we know $\boldsymbol{\theta}(0)$. In other words, we can consider $\boldsymbol{\theta}(t_1^*)$ as two distinct functions, $\bar{\boldsymbol{\theta}}=(\bar{a}, \boldsymbol{\omega})$, with $\boldsymbol{\theta}(0)$ as their input. This implies that $\bar{a}(\boldsymbol{\theta}(t_0))=a(t_1^*)$ and $\bar{\boldsymbol{\omega}}(\boldsymbol{\theta}(0))=\boldsymbol{\omega}(t_1^*)$.

\begin{lemma}\label{matrice sub2}
    Suppose that $\boldsymbol{\omega}:=\boldsymbol{\omega}(0) \sim N\left(0, \boldsymbol{I}_d\right), a=a(0) \sim N(0,1)$ and given $\boldsymbol{x}_i, \boldsymbol{x}_j \in \Omega$. If $$m \geq \max\left\{\frac{16 n^2 d^2 C_{\psi, d}}{C_0 \lambda^2} \log \frac{4 n^2}{\delta},\frac{8n^2d^2  \fR_{S,s_1}\left(\boldsymbol{\theta}(0)\right)}{(\lambda_{a,1}+\lambda_{\boldsymbol{\omega},1})^2}\right\}$$ then with probability at least $1-\delta$ over the choice of $\boldsymbol{\theta}(0)$, we have
    
(i) if $\mathrm{X}:=\sigma_{s_2}\left(\bar{\boldsymbol{\omega}}^{\top}(\boldsymbol{\omega}) \boldsymbol{x}_i\right) \sigma_{s_2}\left(\bar{\boldsymbol{\omega}}^{\top}(\boldsymbol{\omega})\cdot \boldsymbol{x}_j\right)$, then $\|\mathrm{X}\|_{\psi_1} \leq 2d C_{\psi, d}+ \frac{2d^2\psi(m)^2}{\log 2}$.

(ii) if $\mathrm{X}:=\bar{a}(a)^2 \sigma_{s_2}^{\prime}\left(\bar{\boldsymbol{\omega}}^{\top}(\boldsymbol{\omega}) \boldsymbol{x}_i\right) \sigma_{s_2}^{\prime}\left(\bar{\boldsymbol{\omega}}^{\top}(\boldsymbol{\omega}) \boldsymbol{x}_j\right) \boldsymbol{x}_i \cdot \boldsymbol{x}_j$, then $\|\mathrm{X}\|_{\psi_1} \leq 2d C_{\psi, d}+ \frac{2d^2\psi(m)^2}{\log 2}$, where 
\begin{equation}
    \| \mathrm{X} \|_{\psi_1} := \inf\{ s > 0 \mid \mathbf{E}_X[e^{|X|/s}] \leq 2 \},
\end{equation}
\( C_{\psi,d} := \|\chi^2(d)\|_{\psi_1} \) for \( \chi^2(d) \) being the chi-square random variable in \( d \)-dimensional spaces, and where $C_0$ is an absolute constant shown in Proposition \ref{vershynin}.
\end{lemma}
\begin{proof}
    (i) $$|\mathrm{X}| \leq d\|\bar{\boldsymbol{\omega}}(\boldsymbol{\omega})\|_2^2\le 2d \|\boldsymbol{\omega}\|_2^2+2d\|\bar{\boldsymbol{\omega}}(\boldsymbol{\omega})-\boldsymbol{\omega}\|_2^2\le 2d|Z|+2d^2\psi(m)^2$$ where $|Z|:=\|\boldsymbol{\omega}\|_2^2$, and
$$
\begin{aligned}
\|\mathrm{X}\|_{\psi_1} & =\inf \left\{s>0 \mid \mathbf{E}_{\mathrm{X}} \exp (|\mathrm{X}| / s) \leq 2\right\} \\
& =\inf \left\{s>0 \mid \mathbf{E}_{\boldsymbol{w}} \exp \left(\left|\sigma_{s_2}\left(\bar{\boldsymbol{\omega}}^{\top}(\boldsymbol{\omega}) \boldsymbol{x}_i\right) \sigma_{s_2}\left(\bar{\boldsymbol{\omega}}^{\top}(\boldsymbol{\omega})\cdot \boldsymbol{x}_j\right)\right| / s\right) \leq 2\right\} \\
& \leq \inf \left\{s>0 \mid \mathbf{E}_{\boldsymbol{w}} \exp \left(\frac{2d|Z|+2d^2\psi(m)^2}{s}\right) \leq 2\right\} \\
& \le\inf \left\{s>0 \mid \mathbf{E}_{\mathrm{Z}} \exp (2d|\mathrm{Z}| / s) \leq 2\right\}+\inf \left\{s>0 \mid \mathbf{E}_{\boldsymbol{w}} \exp \left(\frac{2d^2\psi(m)^2}{s}\right) \leq 2\right\} \\
& =2d\left\|\chi^2(d)\right\|_{\psi_1}+ \frac{2d^2\psi(m)^2}{\log 2}\\
& \leq 2d C_{\psi, d}+ \frac{2d^2\psi(m)^2}{\log 2}.
\end{aligned}
$$
(ii) $|\mathrm{X}| \leq d|a|^2 \leq 2d|Z|+2d^2\psi(m)^2$ and $\|\mathrm{X}\|_{\psi_1} \leq 2d C_{\psi, d}+\frac{2d^2\psi(m)^2}{\log 2}$.
\end{proof}

To enhance simplicity and maintain consistent notation, we define:
\begin{align}
C_{\psi, d,2} := 2C_{\psi, d} + \frac{2d\psi(m)^2}{\log 2}.
\end{align}

Based on the above lemma and the sub-exponential Bernstein's inequality as outlined in \cite{vershynin2018high} (which we demonstrate in the appendix), we can conclude that the smallest eigenvalue at the beginning of the second iteration remains positive.

\begin{proposition}\label{eig pos2}
    Given $\delta \in(0,1)$ and the sample set $S=\left\{\left(\boldsymbol{x}_i, y_i\right)\right\}_{i=1}^n $ with $\boldsymbol{x}_i$ 's drawn i.i.d. with uniformly distributed. Suppose that Assumption \ref{positive}  holds. If $$m \geq \max\left\{\frac{16 n^2 d^2 C_{\psi, d}}{C_0 \lambda^2} \log \frac{4 n^2}{\delta},n^4\left(\frac{256\sqrt{2} d  \sqrt{\fR_{S,s_1}\left(\boldsymbol{\theta}(0)\right)}}{(\lambda_{a,1}+\lambda_{\boldsymbol{\omega},1})\min\{\lambda_{a,2},\lambda_{\boldsymbol{\omega},2}\}}\log\frac{4m(d+1)}{\delta}\right)\right\}$$ then with probability at least $1-\delta$ over the choice of $\boldsymbol{\theta}(0)$, we have
$$
\lambda_{\min }\left(\boldsymbol{G}_2\left(\boldsymbol{\theta}(t^*_1)\right)\right) \geq\frac{3}{4}(\lambda_{a,2}+\lambda_{2,\boldsymbol{\omega}}),
$$where $C_0$ is an absolute constant shown in Proposition \ref{vershynin}.
\end{proposition}
\begin{proof}
    \begin{align}
    \bar{k}^{[a]}_2(\boldsymbol{x},\boldsymbol{x}'):=&\mathbf{E}_{\boldsymbol{\omega}}\sigma_{s_2}\left(\bar{\boldsymbol{\omega}}^{\top}(\boldsymbol{\omega}) \boldsymbol{x}\right) \sigma_{s_2}\left(\bar{\boldsymbol{\omega}}^{\top}(\boldsymbol{\omega})\cdot \boldsymbol{x}'\right)\notag\\\bar{k}^{[\boldsymbol{\omega}]}_2(\boldsymbol{x},\boldsymbol{x}'):=&\mathbf{E}_{(a,\boldsymbol{\omega})}\bar{a}(a)^2\sigma'_{s_2}\left(\bar{\boldsymbol{\omega}}^{\top}(\boldsymbol{\omega}) \boldsymbol{x}\right) \sigma'_{s_2}\left(\bar{\boldsymbol{\omega}}^{\top}(\boldsymbol{\omega})\boldsymbol{x}'\right)\boldsymbol{x}\cdot\boldsymbol{x}'.
\end{align}

The Gram matrices, denoted as $\bar{\boldsymbol{K}}^{[a]}_2$ and $\bar{\boldsymbol{K}}^{[\boldsymbol{\omega}]}_2$, corresponding to an infinite-width two-layer network with the activation function $\sigma_{s_2}$, can be expressed as follows:\begin{align}
    &\bar{K}_{ij,2}^{[a]}=\bar{k}^{[a]}_2(\boldsymbol{x}_i,\boldsymbol{x}_j),~\bar{\boldsymbol{K}}^{[a]}_2=(\bar{K}_{ij,2}^{[a]})_{n\times n},\notag\\& \bar{K}_{ij,2}^{[\boldsymbol{\omega}]}=\bar{k}^{[\boldsymbol{\omega}]}_2(\boldsymbol{x}_i,\boldsymbol{x}_j),~\bar{\boldsymbol{K}}^{[\boldsymbol{\omega}]}_p=(\bar{K}_{ij,2}^{[\boldsymbol{\omega}]})_{n\times n}.
\end{align}

The proof can be divided into two main parts. The first part, seeks to establish that the difference between $\boldsymbol{K}^{[a]}_2+\boldsymbol{K}^{[\boldsymbol{\omega}]}_2$ and $\bar{\boldsymbol{K}}^{[a]}_2+\bar{\boldsymbol{K}}^{[\boldsymbol{\omega}]}_2$ is small. In this case, the proof draws upon Proposition \ref{bound dy}, which underscores the potential for the error in $\|\boldsymbol{\theta}(0)-\boldsymbol{\theta}(t^*)\|_\infty$ to be highly negligible when $m$ assumes a large value. The second part aims to demonstrate that the disparity between $\boldsymbol{G}(\boldsymbol{\theta}(t_1^*))$ and $\bar{\boldsymbol{K}}^{[a]}_2+\bar{\boldsymbol{K}}^{[\boldsymbol{\omega}]}_2$ is minimal. This particular proof relies on the application of sub-exponential Bernstein's inequality as outlined in \cite{vershynin2018high} (Proposition \ref{vershynin}).

First of all, we prove that the difference between $\boldsymbol{K}^{[a]}_2+\boldsymbol{K}^{[\boldsymbol{\omega}]}_2$ and $\bar{\boldsymbol{K}}^{[a]}_2+\bar{\boldsymbol{K}}^{[\boldsymbol{\omega}]}_2$ is small. Due to \begin{align}
    \left|\bar{k}^{[a]}_2(\boldsymbol{x},\boldsymbol{x}')-k^{[a]}_2(\boldsymbol{x},\boldsymbol{x}')\right|\le&\mathbf{E}_{\boldsymbol{\omega}}\left|\sigma_{s_2}\left(\bar{\boldsymbol{\omega}}^{\top}(\boldsymbol{\omega}) \boldsymbol{x}\right) \sigma_{s_2}\left(\bar{\boldsymbol{\omega}}^{\top}(\boldsymbol{\omega})\boldsymbol{x}'\right)-\sigma_{s_2}\left(\boldsymbol{\omega} \boldsymbol{x}\right) \sigma_{s_2}\left(\boldsymbol{\omega}\cdot \boldsymbol{x}'\right)\right|\notag\\\le&2d\|\bar{\boldsymbol{\omega}}^{\top}(\boldsymbol{\omega}(0))-\boldsymbol{\omega}(0)\|_\infty\|\boldsymbol{\omega}(0)\|_\infty\notag\\\le &2d\psi(m)\sqrt{2\log\frac{4m(d+1)}{\delta}}
\end{align}with probability at least $1-\delta$ over the choice of $\boldsymbol{\theta}(0)$, where the last inequality is due to Lemma \ref{bound initial}. Therefore, \begin{equation}
    \|\boldsymbol{K}^{[a]}_2-\bar{\boldsymbol{K}}^{[a]}_2\|_F\le 2n\psi(m)\sqrt{2\log\frac{4m(d+1)}{\delta}}.
\end{equation} Similarly, we can obtain that \begin{equation}
    \|\boldsymbol{K}^{[\boldsymbol{\omega}]}_2-\bar{\boldsymbol{K}}^{[\boldsymbol{\omega}]}_2\|_F\le 2n\psi(m)\sqrt{2\log\frac{4m(d+1)}{\delta}}.
    \end{equation}

    Set $\psi(m)\le \frac{\min\{\lambda_{a,2},\lambda_{\boldsymbol{\omega},2}\}}{16n\sqrt{2\log\frac{4m(d+1)}{\delta}}}$, i.e. \[m\ge n^4\left(\frac{128\sqrt{2} d  \sqrt{\fR_{S,s_1}\left(\boldsymbol{\theta}(0)\right)}}{(\lambda_{a,1}+\lambda_{\boldsymbol{\omega},1})\min\{\lambda_{a,2},\lambda_{\boldsymbol{\omega},2}\}}2\log\frac{4m(d+1)}{\delta}\right),\] we have \[\max\{\|\boldsymbol{K}^{[a]}_2-\bar{\boldsymbol{K}}^{[a]}_2\|_F,\|\boldsymbol{K}^{[\boldsymbol{\omega}]}_2-\bar{\boldsymbol{K}}^{[\boldsymbol{\omega}]}_2\|_F\}\le \frac{1}{8}\min\{\lambda_{a,2},\lambda_{\boldsymbol{\omega},2}\}.\]

    Furthermore, by sub-exponential Bernstein's inequality as outlined in \cite{vershynin2018high} (Proposition \ref{vershynin}), for any $\varepsilon>0$, we define \begin{align}
        \mathbb{A}_{ij,2}^{[a]}&:=\left\{\boldsymbol{\theta}(0)\mid \left|G_{ij,2}^{[a]}(\boldsymbol{\theta}(0))-\bar{K}_{ij,2}^{[a]}\right|\le\frac{\varepsilon}{n}\right\}\notag\\\mathbb{A}_{ij,2}^{[\boldsymbol{\omega}]}&:=\left\{\boldsymbol{\theta}(0)\mid \left|G_{ij,2}^{[\boldsymbol{\omega}]}(\boldsymbol{\theta}(0))-\bar{K}_{ij,2}^{[\boldsymbol{\omega}]}\right|\le\frac{\varepsilon}{n}\right\}.
    \end{align}

    Setting $\varepsilon\le ndC_{\psi,d,2}$, by Proposition \ref{vershynin} and Lemma \ref{eig pos2}, we have \begin{align}
        \mathbf{P}(\mathbb{A}_{ij,2}^{[a]})&\ge 1-2\exp\left(-\frac{mC_0\varepsilon^2}{n^2d^2C_{\psi,d,2}}\right),\notag\\\mathbf{P}(\mathbb{A}_{ij,2}^{[\boldsymbol{\omega}]})&\ge 1-2\exp\left(-\frac{mC_0\varepsilon^2}{n^2d^2C_{\psi,d,2}}\right).
    \end{align}

    Therefore, with probability at least \[\left[1-2\exp\left(-\frac{mC_0\varepsilon^2}{n^2d^2C_{\psi,d,2}^2}\right)\right]^{2n^2}\ge 1-4n^2\exp\left(-\frac{mC_0\varepsilon^2}{n^2d^2C_{\psi,d,2}^2}\right)\] over the choice of $\boldsymbol{\theta}(0)$, we have \begin{align}
        &\left\|G_2^{[a]}(\boldsymbol{\theta}(0))-\bar{K}_2^{[a]}\right\|_F\le \varepsilon\notag\\&\left\|G_2^{[p]}(\boldsymbol{\theta}(0))-\bar{K}_2^{[p]}\right\|_F\le \varepsilon.
    \end{align}

    Hence by taking $\varepsilon=\frac{1}{8}\min\{\lambda_{a,2},\lambda_{\boldsymbol{\omega},2}\}$ and $\delta=4n^2\exp\left(-\frac{mC_0\lambda_1^2}{16n^2d^2C_{\psi,d,2}^2}\right)$, we obtain that
    \begin{align}
        \lambda_{\min }\left(\boldsymbol{G}_2\left(\boldsymbol{\theta}(t_1^*)\right)\right)\ge &\lambda_{\min }\left(\boldsymbol{G}_2^{[a]}\left(\boldsymbol{\theta}(t_1^*)\right)\right)+\lambda_{\min }\left(\boldsymbol{G}_2^{[\boldsymbol{\omega}]}\left(\boldsymbol{\theta}(t_1^*)\right)\right)\notag\\\ge& \lambda_{a,1}+\lambda_{\boldsymbol{\omega},1}-\left\|\boldsymbol{G}_{2}^{[a]}(\boldsymbol{\theta}(t_1^*)-\bar{\boldsymbol{K}}_{2}^{[a]}\right\|_F-\left\|\boldsymbol{G}_{2}^{[\boldsymbol{\omega}]}(\boldsymbol{\theta}(t_1^*))-\bar{\boldsymbol{K}}_{2}^{[\boldsymbol{\omega}]}\right\|_F\notag\\&-\|\boldsymbol{K}^{[a]}_2-\bar{\boldsymbol{K}}^{[a]}_2\|_F-\|\boldsymbol{K}^{[\boldsymbol{\omega}]}_2-\bar{\boldsymbol{K}}^{[\boldsymbol{\omega}]}_2\|_F\notag\\\ge&\frac{3}{4}(\lambda_{a,2}+\lambda_{\boldsymbol{\omega},2}).
    \end{align}
\end{proof}
\begin{remark}
    In accordance with Proposition \ref{eig pos2}, we can establish that $m$ follows a trend of $\fO\left(\frac{\log(1/\delta)}{\min\{\lambda_{a,2},\lambda_{\boldsymbol{\omega},2}\}}\right)$. This observation sheds light on our strategy of increasing the parameter $s$ with each iteration. As we have proven in Theorem \ref{large}, the smallest eigenvalues of Gram matrices tend to increase as $s$ increases. This insight provides a reason for increasing the activation parameters, thereby enhancing the probability that the Gram matrix is positive and making the energy landscape in this iteration more convex. Now, consider the second iteration. For a fixed value of $m$ that we have at this stage, a larger smallest eigenvalue implies that we can select a smaller value for $\delta$. Consequently, this leads to a higher probability of $\lambda_{\min }\left(\boldsymbol{G}_2\left(\boldsymbol{\theta}(t^*_1)\right)\right)$ being positive.
    
    If we maintain the activation function unchanged and proceed with the proof of Proposition \ref{eig pos2}, it can be inferred that the dynamics at the onset of the second iteration are positive if \( m \) satisfies 
\[
C\frac{\log(1/\delta)}{\min\{\lambda_{a,2},\lambda_{\boldsymbol{\omega},2}\}} \leq m \leq C\frac{\log(1/\delta)}{\min\{\lambda_{a,1},\lambda_{\boldsymbol{\omega},1}\}},
\]
where \( C \) is a constant. However, this positivity guarantee would not hold if the activation functions remain unchanged.
\end{remark}

Set \begin{equation}t^*_2=\inf\{t\mid\boldsymbol{\theta}(t)\not\in\fN_2(\boldsymbol{\theta}(t_1^*))\}\label{t_2}\end{equation} where \[\fN_2(\boldsymbol{\theta}(t_1^*)):=\left\{\boldsymbol{\theta}\mid\|\boldsymbol{G}_2(\boldsymbol{\theta})-\boldsymbol{G}_2(\boldsymbol{\theta}(t_1^*))\|_F\le \frac{1}{4}(\lambda_{a,2}+\lambda_{\boldsymbol{\omega},2})\right\}.\]
\begin{proposition}\label{convergence 2}
    Given the sample set $S=\left\{\left(\boldsymbol{x}_i, y_i\right)\right\}_{i=1}^n \subset \Omega$ with $\boldsymbol{x}_i$ 's drawn i.i.d. with uniformly distributed and $\delta \in(0,1)$. Suppose that Assumption \ref{positive}  holds. $$m \geq \max\left\{\frac{16 n^2 d^2 C_{\psi, d}}{C_0 \lambda^2} \log \frac{4 n^2}{\delta},n^4\left(\frac{256\sqrt{2} d  \sqrt{\fR_{S,s_1}\left(\boldsymbol{\theta}(0)\right)}}{(\lambda_{a,1}+\lambda_{\boldsymbol{\omega},1})\min\{\lambda_{a,2},\lambda_{\boldsymbol{\omega},2}\}}\log\frac{4m(d+1)}{\delta}\right)\right\}$$ then with probability at least $1-\delta$ over the choice of $\boldsymbol{\theta}(0)$, we have for any $t\in[t^*_1,t^*_2]$
\begin{equation}
    \fR_{S,s_2}(\boldsymbol{\theta}(t))\le \fR_{S,s_2}(\boldsymbol{\theta}(t_1^*))\exp\left(-\frac{t-t_1^*}{n} (\lambda_{a,2}+\lambda_{\boldsymbol{\omega},2})\right),
\end{equation}where $C_0$ is an absolute constant shown in Proposition \ref{vershynin}.
\end{proposition}
\begin{proof}
    Due to Proposition \ref{eig pos2} and the definition of $t^*_2$, we have that for any $\delta\in(0,1)$ \begin{equation}
        \lambda_{\min }\left(\boldsymbol{G}_2\left(\boldsymbol{\theta}(t)\right)\right) \geq\frac{1}{2}(\lambda_{a,1}+\lambda_{\boldsymbol{\omega},1})
    \end{equation} for any $t\in[t^*_1,t^*_2]$ with probability at least $1-\delta$ over the choice of $\boldsymbol{\theta}(0)$.

    Then we get that\begin{align}
    \frac{\mathrm{d}}{\mathrm{d}t}\fR_{S,s_2}(\boldsymbol{\theta}(t))=&\sum_{k=1}^m\left(\nabla_{a_k}\fR_{S,s_2}(\boldsymbol{\theta})\frac{\mathrm{d}a_k(t)}{\mathrm{d}t}+\nabla_{\boldsymbol{\omega}_k}\fR_{S,s_2}(\boldsymbol{\theta})\frac{\mathrm{d}\boldsymbol{\omega}_k(t)}{\mathrm{d}t}\right)\notag\\=&-\frac{1}{n^2}\boldsymbol{e}^T_2\boldsymbol{G}_{2}(\boldsymbol{\theta}(t))\boldsymbol{e}_2\notag\\\le&-\frac{2}{n} \lambda_{\min }\left(\boldsymbol{G}_2\left(\boldsymbol{\theta}\right)\right)\fR_{S,s_2}(\boldsymbol{\theta}(t))\notag\\\le&-\frac{1}{n} (\lambda_{a,2}+\lambda_{\boldsymbol{\omega},2})\fR_{S,s_2}(\boldsymbol{\theta}(t)).
\end{align} Therefore, \begin{equation}
    \fR_{S,s_2}(\boldsymbol{\theta}(t))\le \fR_{S,s_2}(\boldsymbol{\theta}(t_1^*))\exp\left(-\frac{t-t_1^*}{n} (\lambda_{a,2}+\lambda_{\boldsymbol{\omega},2})\right).
\end{equation}
\end{proof}

\subsection{Convergence of HRTA}

By combining Propositions \ref{convergence 1} and \ref{convergence 2}, we can establish the convergence of the HRTA.

\begin{theorem}\label{convergence}
    Given $\delta \in(0,1)$, $s_1\in(0,+\infty)$, $\zeta>1$ and the sample set $S=\left\{\left(\boldsymbol{x}_i, y_i\right)\right\}_{i=1}^n$ with $\boldsymbol{x}_i$ 's drawn i.i.d. with uniformly distributed. Suppose that Assumption \ref{positive}  holds, $$m \geq \max\left\{\frac{16 n^2 d^2 C_{\psi, d}}{C_0 \lambda^2} \log \frac{4 n^2}{\delta},n^4\left(\frac{256\sqrt{2} d  \sqrt{\fR_{S,s_1}\left(\boldsymbol{\theta}(0)\right)}}{(\lambda_{a,1}+\lambda_{\boldsymbol{\omega},1})\min\{\lambda_{a,2},\lambda_{\boldsymbol{\omega},2}\}}\log\frac{4m(d+1)}{\delta}\right)\right\}$$ then with probability at least $1-\delta$ over the choice of $\boldsymbol{\theta}(0)$, we have
$$\left\{\begin{array}{l}
\fR_{S,s_1}(\boldsymbol{\theta}(t))\le \fR_{S,s_1}(\boldsymbol{\theta}(0))\exp\left(-\frac{t}{n} (\lambda_{a,1}+\lambda_{\boldsymbol{\omega},1})\right),t\in[0,t_1^*] \\
\fR_{S,s_2}(\boldsymbol{\theta}(t))\le \fR_{S,s_2}(\boldsymbol{\theta}(t_1^*))\exp\left(-\frac{t-t_1^*}{n} (\lambda_{a,2}+\lambda_{\boldsymbol{\omega},2})\right),~t\in[t^*_1,t^*_2].
\end{array}\right.
$$ where $t_i^*$ are defined in Eqs.~(\ref{t_1},\ref{t_2}). Furthermore, we have that the decay speed in $[t_1^*, t_2^*]$ can be faster than $[0,t_1^*]$, i.e. \begin{align}\fR_{S,s_2}(\boldsymbol{\theta}(t))\le &\fR_{S,s_2}(\boldsymbol{\theta}(t_1^*))\exp\left(-\frac{t-t_1^*}{n} (\lambda_{a,2}+\lambda_{\boldsymbol{\omega},2})\right)\notag\\\le& \fR_{S,s_2}(\boldsymbol{\theta}(t_1^*))\exp\left(-\frac{t-t_1^*}{n} (\lambda_{a,1}+\lambda_{\boldsymbol{\omega},1})\right),\end{align}where $C_0$ is an absolute constant shown in Proposition \ref{vershynin}.
\end{theorem}
\begin{proof}
   By amalgamating Propositions \ref{convergence 1} and \ref{convergence 2}, we can readily derive the proof for Theorem \ref{convergence}.
\end{proof}

\begin{corollary} Given $\delta \in(0,1)$, $s_1\in(0,+\infty)$, $\zeta>1$ and the sample set $S=\left\{\left(\boldsymbol{x}_i, y_i\right)\right\}_{i=1}^n$ with $\boldsymbol{x}_i$ 's drawn i.i.d. with uniformly distributed. Suppose that Assumption \ref{positive}  holds, $$m \geq \max\left\{\frac{16 n^2 d^2 C_{\psi, d}}{C_0 \lambda^2} \log \frac{4 n^2}{\delta},n^4\left(\frac{256\sqrt{2} d  \sqrt{\fR_{S,s_1}\left(\boldsymbol{\theta}(0)\right)}}{(\lambda_{a,1}+\lambda_{\boldsymbol{\omega},1})\min\{\lambda_{a,2},\lambda_{\boldsymbol{\omega},2}\}}\log\frac{4m(d+1)}{\delta}\right)\right\}$$ and \(s_2 := \inf\left\{s \in (s_1, 2) \mid \fR_{S,s}\left(\boldsymbol{\theta}(t_1^*)\right) > \zeta \fR_{S,s_1}\left(\boldsymbol{\theta}(t_1^*)\right)\right\},\) then with probability at least $1-\delta$ over the choice of $\boldsymbol{\theta}(0)$, we have for any $t\in[t_1^*,t_2^*]$ \begin{equation}
    \fR_{S,s_2}(\boldsymbol{\theta}(t))\le \zeta\fR_{S,s_1}(\boldsymbol{\theta}(0))\exp\left(-\frac{t_1^*}{n} (\lambda_{a,1}+\lambda_{\boldsymbol{\omega},1})\right)\exp\left(-\frac{t-t_1^*}{n} (\lambda_{a,2}+\lambda_{\boldsymbol{\omega},2})\right).
\end{equation}
\end{corollary}
\begin{remark}
    In the case where $M\ge 2$, note that we may consider the scenario where $m$ becomes larger. However, it's important to emphasize that the order of $m$ remains at $\fO(n^4)$. This order does not increase substantially due to the fact that all the derivations presented in Subsection \ref{t2 it} can be smoothly generalized.
\end{remark}

Building upon the proof of Theorem \ref{convergence}, we can recognize the advantages of the HRTA. In the initial iteration, the training process does not differ significantly from training using training for pure ReLU networks. The traditional method can effectively reduce the loss function within the set $\fN_1(\boldsymbol{\theta}(0))$, defined as:\[\fN_1(\boldsymbol{\theta}(0)):=\left\{\boldsymbol{\theta}\mid\|\boldsymbol{G}_1(\boldsymbol{\theta})-\boldsymbol{G}_1(\boldsymbol{\theta}(0))\|_F\le \frac{1}{4}(\lambda_{a,1}+\lambda_{\boldsymbol{\omega},1})\right\}.\]In other words, the traditional method can effectively minimize the loss function within the time interval $[0,t_1^*]$, where \(t^*_1=\inf\{t\mid\boldsymbol{\theta}(t)\not\in\fN_1(\boldsymbol{\theta}(0))\}.\)Outside of this range, the training speed may slow down significantly and take a long time to converge. However, the HRTA  transits the training dynamics to a new kernel by introducing a new activation function. This allows training to converge efficiently within a new range of $\boldsymbol{\theta}$, defined as:\[\fN_2(\boldsymbol{\theta}(t_1^*)):=\left\{\boldsymbol{\theta}\mid\|\boldsymbol{G}_2(\boldsymbol{\theta})-\boldsymbol{G}_2(\boldsymbol{\theta}(t_1^*))\|_F\le \frac{1}{4}(\lambda_{a,2}+\lambda_{\boldsymbol{\omega},2})\right\},\] if $\boldsymbol{G}_2(\boldsymbol{\theta}(t_1^*))$ is strictly positive definite. Furthermore, we demonstrate that the minimum eigenvalue of $\boldsymbol{G}_2(\boldsymbol{\theta}(t_1^*))$ surpasses that of $\boldsymbol{G}_1(\boldsymbol{\theta}(0))$ under these conditions, as indicated by Theorem \ref{large}. This implies that, rather than decaying, the training speed may actually increase. This is one of the important reasons why we believe that relaxation surpasses traditional training methods in neural network training. Furthermore, in this paper, we provide evidence that $\mathbf{G}_2(\boldsymbol{\theta}(t_1^*))$ indeed becomes strictly positive definite when the width of neural networks is sufficiently large. Building upon Proposition \ref{eig pos2}, we can see that the increasing smallest eigenvalue of Gram matrices in each iteration contributes to a higher likelihood of $\mathbf{G}_2(\boldsymbol{\theta}(t_1^*))$ becoming strictly positive definite.

As the number of iterations increases, the algorithm may not always be effective. On one hand, the time $t^*_i$ will become smaller and smaller if we do not increase the width of the neural networks. On the other hand, at the beginning of each iteration, the loss functions will increase. If the time of decrease is very small, the loss function may not decrease to a lower value than in the previous iteration. Identifying an appropriate number of iterations and suitable homotopy parameters will be considered as part of future work.

In summary, HRTA offers two key advantages in training:

$\bullet$ It dynamically builds the activation function, allowing loss functions to resume their decay when the training progress slows down, all without compromising the accuracy of the approximation.

$\bullet$ It accelerates the decay rate by increasing the smallest eigenvalue of Gram matrices with each homotopy iteration. Consequently, it enhances the probability of Gram matrices becoming positive definite in each iteration, further improving the training process.
\vspace{0.0in}
\section{Experimental Results for the Homotopy Relaxation Training Algorithm}\label{experiment}\vspace{0.1in}
In this section, we will use several numerical examples to demonstrate our theoretical analysis results.
\vspace{0.1in}\subsection{Function approximation by HRTA}\vspace{0.1in}  % 
In the first part, our objective is to employ NNs to approximate functions of the form $\sin\left(2\pi\sum_{i=1}^d x_i\right)$ for both $d=1$ and $d=3$. We will compare the performance of the HRTA method with the Adam optimizer. We used 100 uniform grid points for $d=1$ and 125 uniform grid points for $d=3$. Additional experiment details are provided in Appendix. The following Figures \ref{1d} and \ref{3d} showcase the results achieved using a two-layer neural network with 1000 nodes to approximate $\sin\left(2\pi\sum_{i=1}^d x_i\right)$ for both $d=1$ and $d=3$. 
\begin{remark}We observe oscillations in the figures, which result from plotting the loss against iterations using a logarithmic scale. To mitigate these fluctuations, we decrease the learning rate, allowing the oscillations to gradually diminish during the later stages of training. It's worth noting that these oscillations occur in both the Adam and HRTA optimization algorithms and do not significantly impact the overall efficiency of HRTA.\end{remark}
\begin{figure}[htbp]\vspace{0.0in}
\centering
\begin{minipage}[t]{0.48\textwidth}
\centering
\includegraphics[width=6cm]{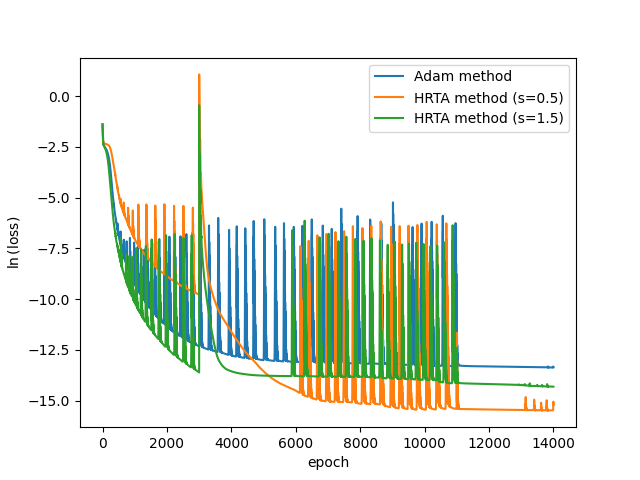}
\caption{Approximation for $\sin(2\pi x)$}\label{1d}
\end{minipage}
\begin{minipage}[t]{0.48\textwidth}
\centering
\includegraphics[width=6cm]{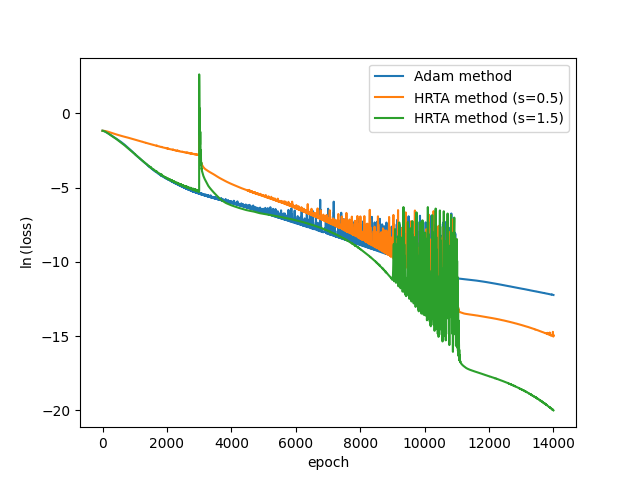}
\caption{Approximation for $\sin(2\pi(x_1+x_2+x_3))$}\label{3d}
\end{minipage}
\end{figure}

In our approach, the HRTA method with $s=0.5$ signifies that we initially employ $\sigma_{\frac{1}{2}}(x):=\frac{1}{2}\text{Id}(x)+\frac{1}{2}\text{ReLU}(x)$ as the activation functions. We transition to using ReLU as the activation function when the loss function does not decay rapidly. This transition characterizes the homotopy part of our method. Conversely, the HRTA kernel with $s=1.5$ signifies that we begin with ReLU as the activation functions and switch to $\sigma_{\frac{3}{2}}(x):=-\frac{1}{2}\text{Id}(x)+\frac{3}{2}\text{ReLU}$ as the activation functions when the loss function does not decay quickly. This transition represents the relaxation part of our method. Based on these experiments, it becomes evident that both cases, $s=0.5$ and $s=1.5$, outperform the traditional method in terms of achieving lower error rates. The primary driver behind this improvement is the provision of two opportunities for the loss function to decrease. While the rate of decay in each step may not be faster than that of the traditional method, as observed in the case of $s=0.5$ for approximating $\sin(2\pi x)$ and $\sin(2\pi(x_1+x_2+x_3))$, it's worth noting that the smallest eigenvalue of the training dynamics is smaller than that of the traditional method when $s=0.5$, as demonstrated in Theorem \ref{large}. This is the reason why, in the first step, it decays slower than the traditional method.

In the homotopy case with $s=0.5$, it can be effectively utilized when we aim to train the ReLU neural network as the final configuration while obtaining a favorable initial value and achieving smaller errors than in the previous stages.
Conversely, in the relaxation case with $s=1.5$, it is valuable when we initially train the ReLU neural network, but the loss function does not decrease as expected. In this situation, changing the activation functions allows the error to start decreasing again without affecting the approximation quality. The advantage of both of these cases lies in their provision of two opportunities and an extended duration for the loss to decrease, which aligns with the results demonstrated in Theorem \ref{convergence}. This approach ensures robust training and improved convergence behavior in various scenarios.

Furthermore, our method demonstrates its versatility as it is not limited to very overparameterized cases or two-layer neural networks. We have shown its effectiveness even in the context of three-layer neural networks and other numbers of nodes (i.e., widths of 200, 400, and 1000). The error rates are summarized in Table \ref{table1}, and our method consistently outperforms traditional methods.
\begin{table}
\centering
\begin{tabular}{|c|c|c|c|c|}
\hline & \multicolumn{2}{|c|}{\text { Single Layer }} & \multicolumn{2}{|c|}{\text { Multi Layer }} \\
\hline & \text { Adam } & \text { HRTA } & \text { Adam } & \text { HRTA } \\
\hline 200 & $2.02\times 10^{-5}$  & $8.72\times 10^{-7}$ (s=1.5)&  $2.52\times 10^{-7}$  & $9.47 \times 10^{-8}$ (s=0.5)\\
\hline 400 & $6.54 \times 10^{-6}$ &  $8.55\times 10^{-7}$ (s=0.5)  &  $4.83\times 10^{-8}$  & $1.8 \times 10^{-8}$ (s=1.5) \\
\hline 1000 & $1.55 \times 10^{-6}$ & $1.88 \times 10^{-7}$ (s=0.5) &  $2.20\times 10^{-7}$  & $3.52 \times 10^{-9}$ (s=1.5) \\
\hline
\end{tabular}\caption{Comparisons between HRTA and Adam methods on different NNs.}\label{table1}
\end{table}\subsection{Solving partial differential equation by HRTA}

In the second part, our goal is to solve the Poisson equation as follows:

\begin{equation}
    	\begin{cases}-\Delta u(x_1,x_2)=f(x_1,x_2)& \text { in } \Omega, \\ \frac{\partial u}{\partial \nu} =0 & \text { on } \partial \Omega,\end{cases}\label{possion}
    \end{equation} 
using HRTA. Here $\Omega$ is a domain within the interval $[0,1]^2$ and \[f(x_1,x_2)=\pi^2\left[\cos(\pi x_1)+\cos(\pi x_2)\right].\] The exact solution to this equation is denoted as $u^*(x_1,x_2) = \cos(\pi x_1) + \cos(\pi x_2)$. We still performed two iterations with $400$ sample points and employed $1000$ nodes. However, we used the activation function $\bar{\sigma}_{\frac{1}{2}}(x)=\frac{1}{2}\text{Id}(x)+\frac{1}{2}\bar{\sigma}(x)$, where $\bar{\sigma}(x)=\frac{1}{2}\text{ReLU}^2(x)$, which is smoother. Here we consider to solve partial differential equations by Deep Ritz method \cite{yu2018deep}. In the deep Ritz method, the loss function of the Eq.~(\ref{possion}) can be read as \[\fE_D(\boldsymbol{\theta}):= \frac{1}{2} \int_{\Omega}|\nabla \phi(\boldsymbol{x};\boldsymbol{\theta})|^2  \mathrm{d}\boldsymbol{x} + \frac{1}{2}\left(\int_{\Omega} \phi(\boldsymbol{x};\boldsymbol{\theta})  \mathrm{d}\boldsymbol{x}\right)^2 - \int_{\Omega} f \phi(\boldsymbol{x};\boldsymbol{\theta})  \mathrm{d}\boldsymbol{x},\]
where $\boldsymbol{\theta}$ represents all the parameters in the neural network. Here, $\Omega$ denotes the domain $[0,1]^d$. Proposition 1 in \cite{lu2021priori} establishes the equivalence between the loss function $\fE_D(\boldsymbol{\theta})$ and $\|\phi(\boldsymbol{x};\boldsymbol{\theta})-u^*(\boldsymbol{x})\|_{H^1([0,1]^2)}$, where $u^*(\boldsymbol{x})$ denotes the exact solution of the PDEs which is $u^*(x_1,x_2) = \cos(\pi x_1) + \cos(\pi x_2)$, and \[\|f\|_{H^1([0,1]^2)}:=\left(\sum_{0 \leq|\alpha| \leq 1}\left\|D^{\boldsymbol{\alpha}} f\right\|_{L^2([0,1]^2)}^p\right)^{1 / 2}.\] Therefore, we can use supervised learning via Sobolev training \cite{czarnecki2017sobolev,son2021sobolev,vlassis2021sobolev} to solve the Poisson equation efficiently and accurately. Our experiments reveal that HRTA remains effective even when $s=1.5$, as demonstrated in Figures \ref{h1} and \ref{l2}: \begin{figure}[htbp]\vspace{0.0in}
\centering
\begin{minipage}[t]{0.48\textwidth}
\centering
\includegraphics[width=6cm]{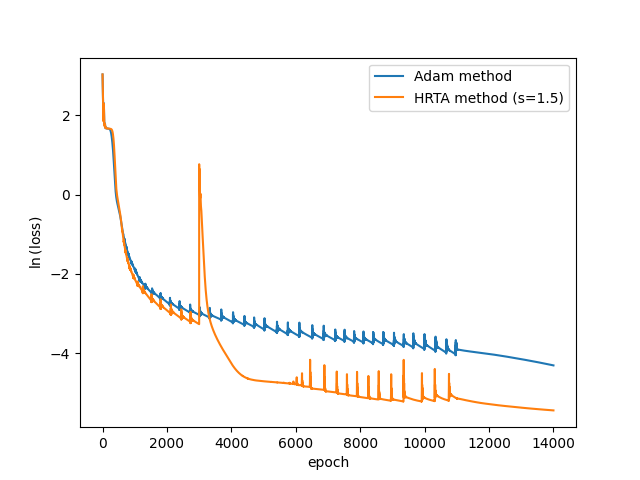}
\caption{Loss function in Deep Ritz method}\label{h1}
\end{minipage}
\begin{minipage}[t]{0.48\textwidth}
\centering
\includegraphics[width=6cm]{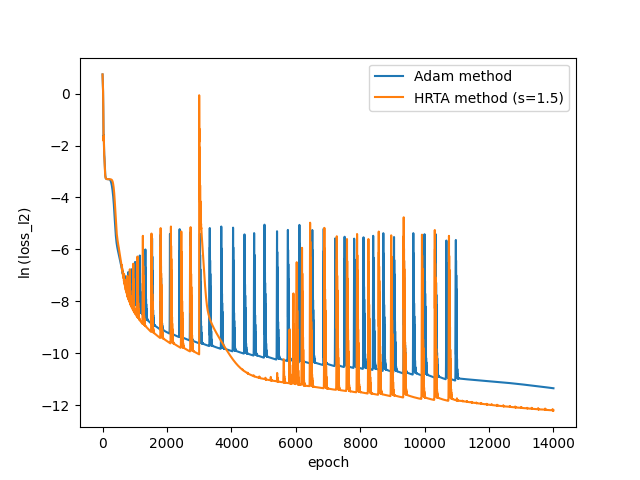}
\caption{Solving Eq.~(\ref{possion}) measured by $L^2$ norm}\label{l2}
\end{minipage}
\end{figure}
\vspace{0.0in}
\section{Conclusion}
In summary, this paper introduces the Homotopy Relaxation Training Algorithm (HRTA), a method designed to expedite gradient descent when it encounters slowdowns. HRTA achieves this by relaxing homotopy activation functions to reshape the energy landscape of loss functions during slow convergence. Specifically, we adapt activation functions to boost the minimum eigenvalue of the gradient descent kernel, thereby accelerating convergence and increasing the likelihood of a positive minimum eigenvalue at each iteration. This paper establishes the theoretical basis for our algorithm, focusing on hyperparameters, while leaving the analysis in a more generalized context for future research. 

\appendix\label{detail}

\section{Sub-exponential Bernstein's Inequality}
\begin{definition}[\cite{vershynin2018high}]
    A random variable $X$ is sub-exponential if and only if its sub-exponential norm is finite i.e.\begin{equation}
        \|X\|_{\psi_1}:=\inf\{s>0\mid\mathbf{E}_X[e^{|X|/s}\le 2]\}.
    \end{equation}
\end{definition}

\begin{proposition}[sub-exponential Bernstein's inequality \cite{vershynin2018high}]\label{vershynin}
    Suppose that $\mathrm{X}_1, \ldots, \mathrm{X}_m$ are i.i.d. sub-exponential random variables with $\mathbf{E} \mathrm{X}_1=\mu$, then for any $s \geq 0$ we have
$$
\mathbf{P}\left(\left|\frac{1}{m} \sum_{k=1}^m \mathrm{X}_k-\mu\right| \geq s\right) \leq 2 \exp \left(-C_0 m \min \left(\frac{s^2}{\left\|\mathrm{X}_1\right\|_{\psi_1}^2}, \frac{s}{\left\|\mathrm{X}_1\right\|_{\psi_1}}\right)\right),
$$
where $C_0$ is an absolute constant.
\end{proposition}

\section{Function approximation using supervised learning}
\begin{example}[Approximating $\sin(2\pi x)$]\label{example 1}
    In the first example, our goal is to approximate the function $\sin(2\pi x)$ within the interval $[0,1]$ using two-layer neural networks (NNs) and the HRTA. We will provide a detailed explanation of the training process for the case of $s=0.5$, which corresponds to the homotopy training case. The training process is divided into two steps:

1. In the first step, we employ the following approximation function:
\begin{equation}
\phi_{\frac{1}{2}}(x;\boldsymbol{\theta}):=\frac{1}{\sqrt{1000}}\sum_{k=1}^{1000} a_k\sigma_{\frac{1}{2}} (\omega_k x)
\end{equation}
to approximate the function $\sin(2\pi x)$. Here, $\sigma_{\frac{1}{2}}(x)=\frac{1}{2}\text{Id}(x)+\frac{1}{2}\sigma(x)$, and the initial values of the parameters are drawn from a normal distribution $\boldsymbol{\theta}\sim\fN(\boldsymbol{0},\boldsymbol{I})$. We select random sample points (or grid points) $\{x_i\}_{i=1}^{100}$, which are uniformly distributed in the interval $[0,1]$. The loss function in this step is defined as \begin{equation}
    \fR_{S,\frac{1}{2}}(\boldsymbol{\theta}):=\frac{1}{200}\sum_{i=1}^{100}|f(x_i)-\phi_{\frac{1}{2}}(x_i;\theta)|^2.
\end{equation} Therefore, we employ the Adam optimizer to train this model over 3000 steps to complete the first step of the process.

%the discrete dynamic system for Gradient Descent (GD) can be expressed as:\[\boldsymbol{\theta}(t+1)=\boldsymbol{\theta}(t)-0.01 \nabla_{\boldsymbol{\theta}}\fR_{S,\frac{1}{2}}(\boldsymbol{\theta}).\] To be more specific, it can be represented as: $$\left\{\begin{array}{l}
%a_k(t+1)=a_k(t)-\frac{1}{200000} \sum_{i=1}^{100}  |f(x_i)-\phi_{\frac{1}{2}}(x_i;\boldsymbol{\theta}_t)|\sigma_{\frac{1}{2}}\left(w_k(t) x_i\right) \\
%\omega_k(t+1)=\omega_k(t)-\frac{1}{200000} \sum_{i=1}^{100}  |f(x_i)-\phi_{\frac{1}{2}}(x_i;\boldsymbol{\theta}_t)|a_k(t)\sigma_{\frac{1}{2}}'\left(w_k(t) x_i\right)x_i
%\end{array}\right.
%$$ and this training process involves iterating these equations for a total of %3000 steps to complete step 1.

2. In the second step, we employ the following approximation function:
\begin{equation}
\phi(x;\boldsymbol{\theta}):=\frac{1}{\sqrt{1000}}\sum_{k=1}^{1000} a_k\sigma (\omega_k x)
\end{equation}
to approximate the function $\sin(2\pi x)$. Here the initial values of the parameters are the results in the first step. The loss function in this step is defined as \begin{equation}
    \fR_{S}(\boldsymbol{\theta}):=\frac{1}{200}\sum_{i=1}^{100}|f(x_i)-\phi(x_i;\boldsymbol{\theta})|^2.
\end{equation} Therefore, we employ the Adam optimizer to train this model over 13000 steps to complete the second step of the process and finish the training.

%can be expressed as:\[\boldsymbol{\theta}(t+1)=\boldsymbol{\theta}(t)-0.01 \nabla_{\boldsymbol{\theta}}\fR_{S}&(\boldsymbol{\theta}).\] To be more specific, it can be represented as: $$\left\{\begin{array}&{l}
%a_k(t+1)=a_k(t)-\frac{1}{200000} \sum_{i=1}^{100}  |f(x_i)-%\phi(x_i;\boldsymbol{\theta}_t)|\sigma\left(w_k(t) x_i\right) \\\omega_k(t+1)=\omega_k(t)-\frac{1}{200000} \sum_{i=1}^{100}  |f(x_i)-\phi(x_i;\boldsymbol{\theta}_t)|a_k(t)\sigma'\left(w_k(t) x_i\right)x_i
%\end{array}\right.
%$$ and this training process involves iterating these equations for a total of 2000 steps to complete step 1.

For the purpose of comparison, we employ a traditional method with the following approximation function:
\begin{equation}
\phi(x;\boldsymbol{\theta}):=\frac{1}{\sqrt{1000}}\sum_{k=1}^{1000} a_k\sigma (\omega_k x)
\end{equation}
to approximate the function $\sin(2\pi x)$. Here, the initial values of the parameters are sampled from a normal distribution $\boldsymbol{\theta} \sim \mathcal{N}(\boldsymbol{0}, \boldsymbol{I})$. We select the same random sample points (or grid points) ${x_i}_{i=1}^{100}$ as used in the HRTA. The loss function in this step is defined as \begin{equation}
    \fR_{S}(\boldsymbol{\theta}):=\frac{1}{200}\sum_{i=1}^{100}|f(x_i)-\phi(x_i;\theta)|^2.
\end{equation} Therefore, we employ the Adam optimizer to train this model over 16000 steps to complete the training.

In addition, we conducted experiments with neural networks that were not highly overparameterized, containing only 200 and 400 nodes. The results are illustrated in the following figures: 
\begin{figure}[htbp]\vspace{0.0in}
\centering
\begin{minipage}[t]{0.48\textwidth}
\centering
\includegraphics[width=6cm]{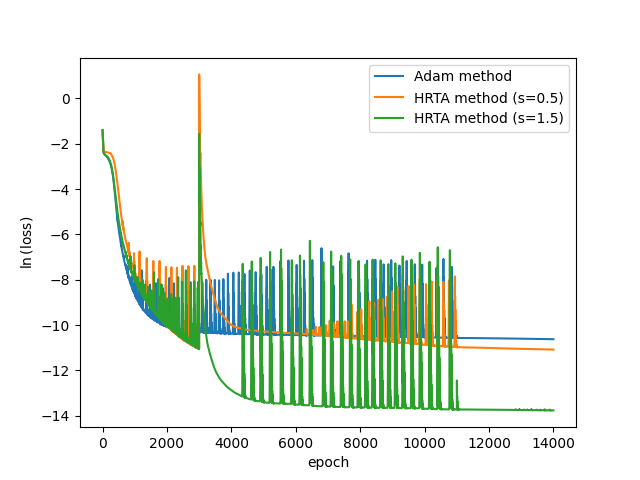}
\caption{Approximation for $\sin(2\pi x)$ by NNs with $200$ nodes}\label{200}
\end{minipage}
\begin{minipage}[t]{0.48\textwidth}
\centering
\includegraphics[width=6cm]{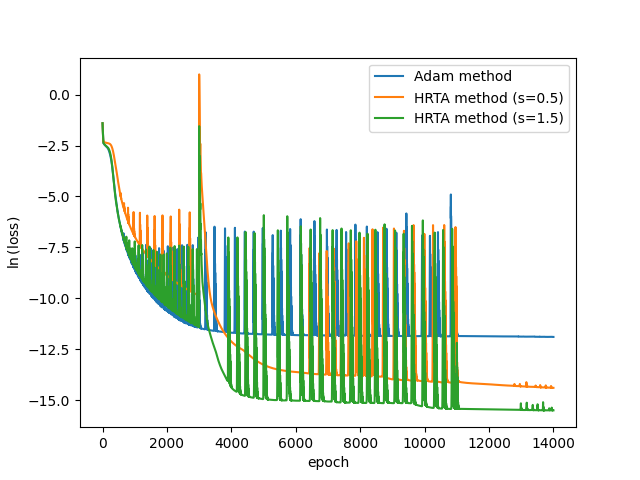}
\caption{Approximation for $\sin(2\pi x)$ by NNs with $400$ nodes}\label{400}
\end{minipage}
\end{figure}
\end{example}

\begin{example}[Approximating $\sin(2\pi(x_1+x_2+x_3))$]
    The training methods in Example \ref{example 1} and this current scenario share the same structure. The only difference is that in this case, all instances of $\omega$ and $x$ used in Example \ref{example 1} have been extended to three dimensions. In Figure \ref{3d}, we demonstrate that HRTA is effective in a highly overparameterized scenario, comprising 125 sample points with 1000 nodes. Additionally, we illustrate that HRTA remains effective in a scenario with less overparameterization, involving 400 nodes and 400 sample points. The results are presented below Figure \ref{400_1000}. \begin{figure}[h!]
	\centering
	\includegraphics[scale=0.60]{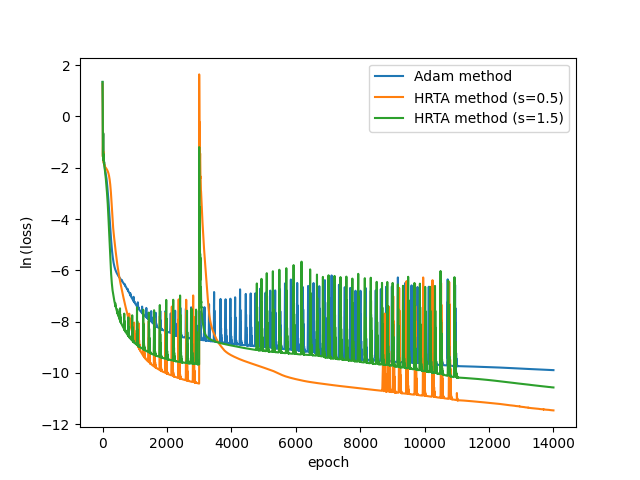}
	\caption{Approximation for $\sin(2\pi(x_1+x_2+x_3))$ with less overparameterization}
	\label{400_1000}
\end{figure}
\end{example}

\section{Solving partial differential equations}
\begin{example}
    In this example, we aim to solve the Poisson equation given by:\begin{equation}
    	\begin{cases}-\Delta u(x_1,x_2)=\pi^2\left[\cos(\pi x_1)+\cos(\pi x_2)\right] & \text { in } \Omega, \\ \frac{\partial u}{\partial \nu} =0 & \text { on } \partial \Omega,\end{cases}\notag
    \end{equation} by homotopy relaxation training methods and Deep Ritz method \cite{yu2018deep}, where $\Omega$ is a domain within the interval $[0,1]^2$. The exact solution to this equation is denoted as $u^*(x_1,x_2) = \cos(\pi x_1) + \cos(\pi x_2)$.

    1. In the first step, we employ the following approximation function:
\begin{equation}
\bar{\phi}(\boldsymbol{x};\boldsymbol{\theta}):=\frac{1}{\sqrt{1000}}\sum_{k=1}^{1000} a_k\bar{\sigma} (\boldsymbol{\omega}_k \boldsymbol{x})
\end{equation}
to solve Passion equations. Here, $\bar{\sigma}(x)=\frac{1}{2}\text{ReLU}^2(x)$, and the initial values of the parameters are drawn from a normal distribution $\boldsymbol{\theta}\sim\fN(\boldsymbol{0},\boldsymbol{I})$. We select random sample points (or grid points) $\{x_i\}_{i=1}^{400}$, which are uniformly distributed in the interval $[0,1]^2$. As per \cite[Proposition 1]{lu2021priori}, the loss function in the Deep Ritz method for solving this Poisson equation is indeed given by: \begin{equation}
    \fR_{S,\frac{1}{2}}(\boldsymbol{\theta}):=\frac{1}{800}\sum_{i=1}^{400}\left[|u^*(\boldsymbol{x}_i)-\bar{\phi}(\boldsymbol{x}_i;\boldsymbol{\theta})|^2+|\nabla u^*(\boldsymbol{x}_i)-\nabla\bar{\phi}(\boldsymbol{x}_i;\boldsymbol{\theta})|^2\right].
\end{equation}This loss function captures the discrepancy between the exact solution $u^*(\boldsymbol{x}_i)$ and the network's output $\bar{\phi}(\boldsymbol{x}_i;\boldsymbol{\theta})$, as well as the gradient of the exact solution and the gradient of the network's output, for each sampled point $\boldsymbol{x}_i$. Therefore, we employ the Adam optimizer to train this model over 16000 steps to complete the step.

2. In the second step, we employ the following approximation function:
\begin{equation}
\bar{\phi}_{\frac{3}{2}}(x;\boldsymbol{\theta}):=\frac{1}{\sqrt{1000}}\sum_{k=1}^{1000} a_k\bar{\sigma} (\omega_k x)
\end{equation}
to solve Possion equations. Here the initial values of the parameters are the results in the first step. The loss function in this step is defined as \begin{equation}
    \fR_{S}(\boldsymbol{\theta}):=\frac{1}{800}\sum_{i=1}^{400}\left[|u^*(\boldsymbol{x}_i)-\bar{\phi}_{\frac{3}{2}}(\boldsymbol{x}_i;\boldsymbol{\theta})|^2+|\nabla u^*(\boldsymbol{x}_i)-\nabla\bar{\phi}_{\frac{3}{2}}(\boldsymbol{x}_i;\boldsymbol{\theta})|^2\right].
\end{equation} Therefore, we employ the Adam optimizer to train this model over 13000 steps to complete the step.
\end{example}

%\begin{acknowledgements}
%\end{acknowledgements}


\bibliographystyle{spmpsci}
\bibliography{references}

%\begin{thebibliography}{}
% etc
%\end{thebibliography}

\end{document}